\documentclass{article}

\usepackage{arxiv}

\usepackage[utf8]{inputenc} 
\usepackage[T1]{fontenc}    
\usepackage{hyperref}       
\usepackage{url}            
\usepackage{booktabs}       
\usepackage{amsfonts,amsmath}       
\usepackage{nicefrac}       
\usepackage{microtype}      
\usepackage{lipsum}
\usepackage{graphicx}
\usepackage{bm}
\graphicspath{ {./}, {./} }
\usepackage{amssymb}
\usepackage{array}
\usepackage{mathtools}
\usepackage{color}
\usepackage{tabularx,booktabs}

\usepackage{algorithm,algorithmicx,algpseudocode}
\usepackage{enumitem}
\definecolor{mblue}{RGB}{57,106,177}
\hypersetup{
    citecolor=blue,
	colorlinks=true,
	linkcolor=blue,
	filecolor=blue,      
	urlcolor=blue
}
\title{Efficient dynamic modal load reconstruction using physics-informed Gaussian processes based on frequency-sparse Fourier basis functions}

\author{
  Gledson Rodrigo Tondo$^{\star}$ \\
  Bauhaus-Universität Weimar\\
  Weimar, Germany \\
  \And
  Igor Kavrakov \\
  University of Cambridge\\
  Cambridge, United Kingdom \\
  \And
  Guido Morgenthal \\
  Bauhaus-Universität Weimar\\
  Weimar, Germany \\
}

\begin{document}
\maketitle
\begin{abstract}
Knowledge of the force time history of a structure is essential to assess its behaviour, ensure safety and maintain reliability. However, direct measurement of external forces is often challenging due to sensor limitations, unknown force characteristics, or inaccessible load points. This paper presents an efficient dynamic load reconstruction method using physics-informed Gaussian processes (GP) based on frequency-sparse Fourier basis functions. The GP's covariance matrices are built using the description of the system dynamics, and the model is trained using structural response measurements. This provides support and interpretability to the machine learning model, in contrast to purely data-driven methods. In addition, the model filters out irrelevant components in the Fourier basis function by leveraging the sparsity of structural responses in the frequency domain, thereby reducing computational complexity during optimization. The trained model for structural responses is then integrated with the differential equation for a harmonic oscillator, creating a probabilistic dynamic load model that predicts load patterns without requiring force data during training. The model's effectiveness is validated through two case studies: a numerical model of a wind-excited 76-story building and an experiment using a physical scale model of the Lilleb{\ae}lt Bridge in Denmark, excited by a servo motor. For both cases, validation of the reconstructed forces is provided using comparison metrics for several signal properties. The developed model holds potential for applications in structural health monitoring, damage prognosis, and load model validation.
\end{abstract}

\section{Introduction}
\label{sec:Introduction}

Reconstructing the time history of forces that a structure has been subjected to is essential for various engineering applications, including structural health monitoring, condition assessment, damage prognosis and load model validation. Direct measurement of external forces is often limited or impractical due to sensor constraints, the unknown nature of the forces, or the inaccessibility of load points for placing transducers. Consequently, indirect force reconstruction methods are commonly employed, relying on measurement data from the structural responses to applied dynamic loads.

Previous efforts to reconstruct dynamic forces from structural responses are well-documented in the literature, with various strategies employed to recover the applied loads~\cite{sanchez2014review,jayalakshmi2018dynamic}. One common approach involves the use of analytical spatial expressions for reconstructing loads on relatively simple structural systems, such as Euler-Bernoulli beams, thin plates, and cylindrical shells~\cite{jiang2008reconstruction,jiang2009reconstruction,djamaa2007reconstruction}. This method is akin to truncated singular value decomposition and its related techniques~\cite{xu1998truncated,chen2017truncated,DYKES201415}, where the structural responses are projected onto a reduced set of spatial modes. Modal superposition methods represent another widely used strategy, where dynamic loads are reconstructed by regressing a set of modal load time histories, under the assumption of a known structural mode shape matrix. This matrix is typically truncated at a certain modal number to simplify the calculations~\cite{law1997moving,li2015generalized}. The implications of such modal truncation on the accuracy of force identification have been thoroughly examined~\cite{vigso2019effect}. In the time domain, impulse response functions have also been successfully applied to dynamic load reconstruction~\cite{amiri2015derivation,li2023principal}. On the other hand, frequency domain-based methods focus on truncating the frequency content to reconstruct forces~(e.g.~\cite{rezayat2016identification,ghaderi2015practical}). These methods typically operate under the premise that only a limited range of frequencies are relevant for accurate load reconstruction. To address the inherent ill-posed nature of the force reconstruction problem, many approaches employ regularization techniques. Commonly used regularization methods include LASSO regression~\cite{QIAO201671,QIAO201672,QIAO201793} and Tikhonov regularization~\cite{amiri2015derivation,li2014load,mao2010experimental,jacquelin2003force,wensong2018fractional}. More advanced regularization techniques have also been developed, leveraging prior knowledge about the spatio-temporal characteristics of the applied loads or employing Bayesian regularization through maximum \textit{a posteriori} estimation~\cite{aucejo2019space,aucejo2019optimal}. Furthermore, extended Kalman filters have been employed as an alternative technique for dynamic load reconstruction, providing a recursive estimation framework capable of handling non-linearities and uncertainties inherent in the system~\cite{liu2021dynamic,feng2020force,petersen2018estimation}. In recent developments, machine learning methods have emerged as a significant tool in force reconstruction tasks, with a focus on artificial neural networks~\cite{liu2022artificial,chen2022feature} and Gaussian process latent force models~\cite{alvarez2013linear,nayek2019gaussian,petersen2022wind}, the latter of which combines the flexibility of Gaussian processes with the physical interpretability similar to state space models. Additionally, physics-informed machine learning models have been introduced, which integrate physical laws and domain knowledge into the learning process, thereby ensuring that predictions remain consistent with and leverage knowledge of the governing principles of structural dynamics. These models have demonstrated enhanced accuracy and robustness, particularly in situations characterized by sparse or noisy data, where conventional data-driven approaches may struggle to generalize effectively~\cite{tondo2023physics, tondo2024bayesian}.

This paper presents a novel, efficient dynamic load reconstruction model using physics-informed Gaussian processes (GP), with a hybrid formulation based both on input data and on the mathematical description of the physical process to be learned. The model aims to overcome poor generalisation or physically inconsistent predictions, common issues in machine learning methods, while providing explainable and interpretable predictions~\cite{cicirello2024physicsenhancedmachinelearningposition}. We begin by constructing a prior GP model for structural deflections, with the covariance matrix derived from Fourier basis functions. By exploring the sparsity in the signal's frequency content, we allow for the selection of specific frequency components to be included in the basis function matrices, similar to frequency truncation methods. In addition, the formulation of the basis functions ground the model in the physical understanding of the underlying process, providing an inductive bias for learning, in contrast to purely data-driven machine learning models (e.g.~\cite{liu2022artificial,tondo2023physics, tondo2024bayesian, chen2022feature}). To extend the model's applicability, we analytically calculate time derivatives of the basis functions, thereby generating additional models for other quantities such as velocities and accelerations. These functions are used to form sets of covariance and cross-covariance matrices, which together create a multi-output GP model that can be optimized to fit the measurement data effectively. A key advantage of this approach is its ability to learn from repeated sets of similar responses, such as when sensors of varying quality are used, and to jointly learn from heterogeneous data sources, like velocity and acceleration measurements, introducing further model biases to the displacement basis functions. During the optimization process, the GP model inherently balances model complexity with data fitting, eliminating the need for an external regularization parameter, which is often required in traditional force reconstruction methods. The basis functions formulation and the frequency domain sparsity property are leveraged during training to reduce the optimisation's computational complexity. Once the model is trained, the basis functions are combined with the differential equation of the harmonic oscillator to produce a physics-informed model for the forces, which is then used to predict the underlying dynamic loads. This provides a model form bias to the typical GP regression framework. In comparison to the popular linear latent force model~\cite{alvarez2013linear}, which relies on Green's solution for the harmonic oscillator to build the force covariance kernels, the method presented in this paper is build directly via the differential equation, allowing for frequency modulation during the training and prediction steps. Importantly, no force data is required during the training phase, as the model relies solely on the measurement data. This feature is particularly valuable because it allows the model to predict rare or extreme events, such as specific traffic scenarios or unusual wind conditions, as long as the structural system remains linear during the period of measurement. Therefore, the model presented herein integrates and extends multiple existing approaches for dynamic force reconstruction, enhancing previous machine learning-based models~\cite{alvarez2013linear,tondo2023physics}, avoiding issues commonly found in regularization-based models~\cite{QIAO201672,amiri2015derivation}, and building upon advantages existent in frequency truncation methods~\cite{rezayat2016identification}.

This paper is organized as follows: Sec.~\ref{sec:GPModel} introduces the force reconstruction problem, defines the prior Gaussian process model for the measurements, explains how to address computational complexity during training, and derives the model and framework for dynamic force predictions based on the response model. In Sec.~\ref{sec:Fundamentalstudy}, the properties of the measurement GP model are explored in depth, examining the effects of prior assumptions on frequency content, noise levels, and the interplay between heterogeneous measurements. Additionally, the impact of signal properties on load prediction quality is evaluated. Sec.~\ref{sec:Applications} demonstrates the application of the developed model in two scenarios: a 76-story benchmark building excited by wind forces and a physically scaled model of the Lilleb{\ae}lt Bridge in Denmark, excited by a dedicated servo motor. Lastly, Sec.~\ref{sec:Conclusions} provides a summary and conclusions of the study, discussing the model's limitations and potential improvements. The implementation of the model can be found in the GitHub repository available at $\href{https://github.com/gledsonrt/PIGPDynamicModalLoadReconstruction}{\mathrm{github.com/gledsonrt/PIGPDynamicModalLoadReconstruction}}$.

\section{Physics-informed GP model}
\label{sec:GPModel}

\subsection{Problem statement and model description}
\label{sec:GPProblemStatement}

\begin{figure}[!t]
    \centering
    \includegraphics[]{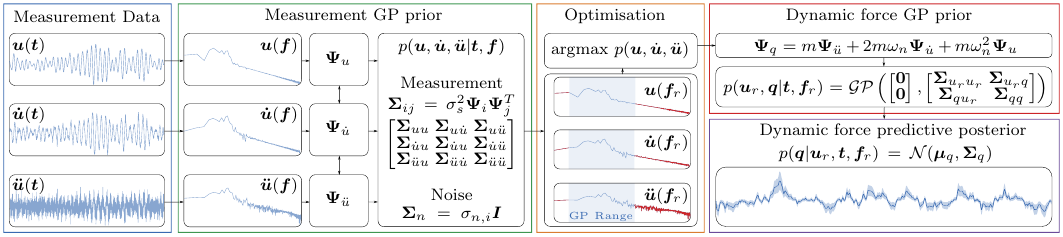}
    \caption{Schematic of the physics-informed Gaussian process model for dynamic load prediction. Using the frequency content of the measurement data (blue box) as inputs, a GP model for the measurements (green box, Sec.~\ref{sec:GPMeasurementPrior}) is created using Fourier basis functions ${\psi}_i(t,f)$ for $i \in \lbrace u, \dot{u}, \Ddot{u}\rbrace$ and a Gaussian noise kernel. The measurement model is reduced and optimised (orange box, Sec.~\ref{sec:GPModelTraining}), by filtering out low-contributing frequency bands and identifying optimal values for the free parameters $\sigma_s$ and $\sigma_{n,i}$. Further, physics-informed basis functions are created for loads (red box, Sec.~\ref{sec:GPModelPrediction}), yielding a joint load-measurement model. Conditioning the model on the measurement data $\bm{u}_r$ yields probabilistic load predictions (violet box, Sec.~\ref{sec:GPModelPrediction}).}\label{fig:PIGP}	
\end{figure}	

The discrete dynamic responses $\bm{z}(\bm{t})$, $\dot{\bm{z}}(\bm{t})$ and $\Ddot{\bm{z}}(\bm{t})$ of a structure due to a dynamic applied force $\bm{p}(\bm{t})$ in the time instants $\bm{t}$ are described by the second order differential equation

\begin{equation}
\bm{M} \Ddot{\bm{z}} + \bm{C} \dot{\bm{z}} + \bm{K} \bm{z} = \bm{p},
\label{eq:dinoscA}
\end{equation}

\noindent where $\bm{M}$, $\bm{C}$ and $\bm{K}$ are the mass, damping and stiffness matrices, respectively. The response signals can be converted into their modal components by pre-multiplication with the structural eigenvector matrix $\bm{\Phi}$, such that e.g. for displacements $\bm{z}~\!\!\!=\!\!\!~\bm{\Phi}~\!\!\bm{u}$, while the same applies for velocities $\dot{\bm{z}}$ and accelerations $\Ddot{\bm{z}}$, and the modal loads are obtained as $\bm{q}~\!\!=~\!\!\bm{\Phi}^T \bm{p}$. Due to eigenvalue orthogonality, the modal mass, damping and stiffness matrices become diagonal and the problem reduces to a set of linearly independent systems, where each mode can be individually solved by

\begin{equation}
m \Ddot{{u}} + 2 m \zeta \omega_n \dot{{u}} + m \omega_n^2 {u} = {q},
\label{eq:dinoscC}
\end{equation}

\noindent where $m$ is the oscillator's mass, $\zeta$ its damping ratio to critical and $\omega_n$ its circular natural frequency. 

In practice, measurements of individual components of the global responses $\bm{z}$, $\dot{\bm{z}}$ and $\Ddot{\bm{z}}$ can be made using various sensing devices (c.f. Fig.~\ref{fig:PIGP}, blue box), and modal decomposition can be done despite an incomplete mode shape matrix using a least squares approach. However, measuring the dynamic load $\bm{p}$ is challenging, often necessitating indirect calculation strategies, such as reconstructing the forcing signal from the responses.

To address this, we first derive a Gaussian process model specifically for the measurable displacements (Fig.~\ref{fig:PIGP}, green box). This model does not require a predefined mean function, and its covariance matrix is constructed using a set of Fourier basis functions, constructed using time $t$ and frequency $f$ components defined by the measurement data. Additionally, the covariance matrix assumes independent and identically distributed (i.i.d.) measurement noise and can distinguish it from the true response through further optimization of model parameters. Connections between different response types (displacements, velocities and accelerations) are created via analytical derivatives, providing a physics-based link between data types (Fig.~\ref{fig:PIGP}, green box, and Sec.~\ref{sec:GPMeasurementPrior}). This approach allows not only for training with - one or more - heterogeneous datasets, where different types of responses are measured, but also for repeated types of datasets, where groups of sensors with varying characteristics measure the same type of response. Optimal model parameters (Fig.~\ref{fig:PIGP}, orange box, and Sec.~\ref{sec:GPModelTraining}) are identified via maximum likelihood optimization, leveraging sparsity in spectral content to filter our frequency bands not contributing to structural response and allowing for an efficient inversion of the covariance matrix.

Once the GP model for measurements is created and optimised, we derive basis functions for the dynamic loads based on the harmonic oscillator equation, shown in Eq.~(\ref{eq:dinoscC}), and construct a joint model for measurements and loads (Fig.~\ref{fig:PIGP}, red box, and Sec.~\ref{sec:GPModelPrediction}). This physics-informed model is then conditioned on the measurement data, providing predictions for the underlying loads characterized by a mean value and a confidence interval (Fig.~\ref{fig:PIGP}, violet box, and Sec.~\ref{sec:GPModelPrediction}).

\subsection{Prior model for structural responses}
\label{sec:GPMeasurementPrior}

We begin by creating prior models for the measurable structural responses. Taking the displacement signal $u(t)$ as a starting point, a Gaussian process prior model can be created as

\begin{equation}
    u(t) \sim \mathcal{GP} \left( 0, k_{uu}(t,t') \right),
    \label{eq:GPdisp}
\end{equation}

\noindent where $k_{uu}(t,t')$ is a covariance kernel that encodes the properties of the displacement signal at time instants $t$ and $t'$. The zero-mean assumption of the prior GP model leads to no loss of generality, as shown in Sec.~\ref{sec:GPModelPrediction}. 

In the GP literature, the covariance kernel $k_{uu}$ is generally modelled by closed-form kernel functions, such as the squared exponential (SE), the rational quadratic or the Matérn family of kernels~\cite{rasmussenGaussianProcessesMachine2006}, (see e.g.~\cite{kavrakov2024data, tondo2023physics, kavrakov2024stochastic}). Herein, the covariance matrix is modelled using a Gramian scheme~\cite{sreeram1994properties}, such that

\begin{equation}
     k_{uu}(t,t',f,f') = \psi_{u}(t,f) \ \psi_{u}(t',f') ,
    \label{eq:GramMat}
\end{equation}

\noindent with $\psi_{u} (t,f)$ being an operator that constructs sets of orthogonal basis functions that can represent the displacement signal. Recalling Fourier's theory, a stationary and infinitely repeating signal can be decomposed into a series of trigonometric functions. We therefore model $\psi_{u}$ as 

\begin{equation}
     \psi_{u}(t, f) = \left[\mathrm{sin}(2 \pi {t}{f}), \mathrm{cos}(2 \pi {t}{f}) \right] {\lambda}_{u} (f) ,
    \label{eq:BasisFunctU}
\end{equation}

\noindent where $f$ is the oscillation frequency, and ${\lambda}_u (f)$ serves as a scale for each basis function $\psi_u(f, {t})$. For the prior model on displacements, we assume $\lambda_u$ as the magnitude of the signal's Fourier transform, scaling both the sine and cosine components as

\begin{equation}
     {\lambda}_u (f) = |\mathcal{F} \left[ {u} \right] (f) |.
    \label{eq:ScalesU}
\end{equation}

\noindent  It is worth noting that the assumptions of Fourier analysis, that the sampled input signal repeats periodically and that the period equals the input length, also apply to the basis functions $\psi_{u}(t, f)$ and therefore to the forcing signals reconstructed by the physics-informed GP model. Modal velocities and accelerations may also be represented as Gaussian processes

\begin{align}
    \dot{{u}} (t) &\sim \mathcal{GP} \left( {0}_, k_{\dot{u}\dot{u}}  (t,t',f,f') \right),\\
    \Ddot{{u}} (t) &\sim \mathcal{GP} \left( {0}, k_{\Ddot{u}\Ddot{u}}  (t,t',f,f') \right),
    \label{eq:GPvelacc}
\end{align}

\noindent where $k_{\dot{u}\dot{u}} = \psi_{\dot{u}} (t,f) \ \psi_{\dot{u}} (t',f')$ and $k_{\Ddot{u}\Ddot{u}} = \psi_{\Ddot{u}} (t,f) \ \psi_{\Ddot{u}} (t',f')$ are the covariance kernels for velocities and accelerations and are generated with specific basis functions, similarly to the displacement case (cf. Eq.~(\ref{eq:GramMat})). They are obtained by leveraging the linear physical relationship between the respective quantities

\begin{align}
     {\psi}_{\dot{u}} &= \frac{d}{dt} \psi_{{u}}({t}, {f}) = 2 \pi f \left[ \mathrm{cos}(2 \pi {t}{f}), -\mathrm{sin}(2 \pi {t}{f}) \right] {\lambda}_{u} (f), \label{eq:BasisVel}\\
     {\psi}_{\Ddot{u}} &= \frac{d^2}{dt^2} \psi_{{u}}({t}, {f}) = 4 \pi^2 f^2 \left[ -\mathrm{sin}(2 \pi {t}{f}), -\mathrm{cos}(2 \pi {t}{f}) \right] {\lambda}_{u} (f).
    \label{eq:BasisAcc}
\end{align}

\noindent It is worth noting that, although we represent ${\psi}_{\dot{u}}$ and ${\psi}_{\Ddot{u}}$ as functions of $\psi_u$, and therefore of ${\lambda}_{u}$, they can also be generated by taking velocities or accelerations as initial points and applying the appropriate linear operators, following the logic in Eqs.~(\ref{eq:BasisVel})~and~(\ref{eq:BasisAcc}). 

Creating the covariance matrices using basis functions also allows for a simple calculation of the cross-covariance between different physical properties, effectively modelling the correlation between heterogeneous data types and the physical interpretation of the time derivative into a covariance matrix. Therefore, in a generalized manner, we calculate the GP's covariance matrices as 

\begin{equation}
     k_{ij} (t,t',f,f') = \psi_i (t,f) \ \psi_j (t',f'),
    \label{eq:GeneralCovCration}
\end{equation}

\noindent for $i,j \in \lbrace u, \dot{u}, \Ddot{u} \rbrace$. 

We now assume a dataset $\mathcal{D} = \lbrace \bm{u}, \dot{\bm{u}}, \ddot{\bm{u}}\rbrace$, which includes the structural modal displacements $\bm{u}$, modal velocities $\dot{\bm{u}}$ and modal accelerations $\ddot{\bm{u}}$, collected simultaneously at $N_t$ time instants $\bm{t}$ with a sampling rate of $f_s = 1/\Delta t$, where $\Delta t$ is the time interval between measurements. Using the developed models, a block-covariance matrix $\bm{\Sigma}_{r}$ that represents the prior assumptions on the structural responses can be created such that 

\begin{equation}
     \bm{\Sigma}_{r} = \bm{\Psi}_{r} \bm{\Psi}_{r}^T = \begin{bmatrix} 
     \bm{\Psi}_{u} \\
     \bm{\Psi}_{\dot{u}} \\
     \bm{\Psi}_{\Ddot{u}} \end{bmatrix} \left[ \bm{\Psi}_{u}, \bm{\Psi}_{\dot{u}}, \bm{\Psi}_{\Ddot{u}} \right] = \begin{bmatrix} 
     \bm{\Sigma}_{uu} & \bm{\Sigma}_{u\dot{u}} & \bm{\Sigma}_{u\Ddot{u}}\\
     \bm{\Sigma}_{\dot{u}u} & \bm{\Sigma}_{\dot{u}u} & \bm{\Sigma}_{\dot{u}\Ddot{u}}\\
     \bm{\Sigma}_{\Ddot{u}u} & \bm{\Sigma}_{\Ddot{u}\dot{u}} & \bm{\Sigma}_{\Ddot{u}\Ddot{u}}\\
     \end{bmatrix},
    \label{eq:CovR}
\end{equation}

\noindent where $\bm{\Psi}_i = \psi_i(\bm{t}, \bm{f})$ for $i \in \lbrace u, \dot{u}, \Ddot{u} \rbrace$, and the response basis functions matrix $\bm{\Psi}_r = [\bm{\Psi}_u, \bm{\Psi}_{\dot{u}}, \bm{\Psi}_{\Ddot{u}}]^T$ has size $3N_t \times 2 N_f$ leading to a covariance matrix $\bm{\Sigma}_r$ of size $3N_t \times 3 N_t$.

In addition to the structural responses modelled by $\bm{\Sigma}_r$, we need to account for the noise content that contaminates the measurement dataset $\mathcal{D}$. Herein, the core assumption is that the noise content is of Gaussian type and is independent and identically distributed (i.i.d.). Taking once again the displacements as a base case, we model the covariance matrix $\bm{\Sigma}_{n,u}$ of the noise content in the displacement data as 

\begin{equation}
     \bm{\Sigma}_{n,u} = \sigma_{n,u}^2 \bm{I},
    \label{eq:NoiseDisp}
\end{equation}

\noindent where $\sigma_{n,u}^2$ is the variance of the noise content in the data, and $\bm{I}$ is the identity matrix of size $N_t \times N_t$. Allowing for individual noise content quantities for velocities and accelerations and leveraging the i.i.d. assumption leads to the full noise model

\begin{equation}
     \bm{\Sigma}_{n} = \begin{bmatrix} 
     \sigma_{n,u}^2 \bm{I} & \bm{0} & \bm{0}\\
     \bm{0} & \sigma_{n,\dot{u}}^2 \bm{I} & \bm{0}\\
     \bm{0} & \bm{0} & \sigma_{n,\Ddot{u}}^2 \bm{I}\\
     \end{bmatrix} = \begin{bmatrix} 
     \bm{\Sigma}_{n,u} & \bm{0} & \bm{0}\\
     \bm{0} & \bm{\Sigma}_{n,\dot{u}} & \bm{0}\\
     \bm{0} & \bm{0} & \bm{\Sigma}_{n,\Ddot{u}}\\
     \end{bmatrix},
    \label{eq:NoiseFull}
\end{equation}

\noindent where $\sigma_{n,\dot{u}}^2$ and $n,\sigma_{\Ddot{u}}^2$  are the variances for the velocity and acceleration dataset noise. The noise covariance matrix has, therefore, the same size as the data covariance matrix, $3 N_t \times 3 N_t$.

Combining the response and noise models and introducing a variance scale $\sigma_s^2$ to control the amplitude of the response w.r.t. the noise yields the full covariance matrix $\bm{\Sigma}$, and the physics-informed Gaussian process prior model for the measurement data takes the form

\begin{equation}
     \left[ \bm{u}, \dot{\bm{u}}, \Ddot{\bm{u}} \right]^T \sim \mathcal{N} \left( \bm{0}, \bm{\Sigma}\right).
    \label{eq:FullGPPriorMeasurements}
\end{equation}

\noindent where $\bm{\Sigma} = \sigma_s^2 \bm{\Sigma}_r + \bm{\Sigma}_n$, with $\bm{\Sigma}_r$ and $\bm{\Sigma}_n$ modelled by Eqs.~(\ref{eq:CovR}) and (\ref{eq:NoiseFull}), respectively.

\subsection{Model reduction and efficient parameter optimisation}
\label{sec:GPModelTraining}

The model derived in Sec. \ref{sec:GPMeasurementPrior} contains free parameters, namely the measurement scale variance $\sigma_s^2$ and the noise variances $\sigma_{n,u}^2$, $\sigma_{n,\dot{u}}^2$ and $\sigma_{n,\Ddot{u}}^2$, that can be identified from the provided training data. Parameter identification in Gaussian process regression is generally carried out via maximum likelihood optimisation~\cite{rasmussenGaussianProcessesMachine2006}:

\begin{equation}
\bm{\theta}_{\mathrm{opt}} = \underset{\bm{\theta}}{\mathrm{argmax}} \ \mathrm{log} \ p(\bm{u}_r|\bm{t},\bm{f},\bm{\theta}) = \underset{\bm{\theta}}{\mathrm{argmax}} \left( - \frac{1}{2} \bm{u}_r^T \bm{\Sigma}^{-1} \bm{u}_r - \frac{1}{2} \mathrm{log} \ \! \mathrm{det} (\bm{\Sigma}) - \frac{3 N_t}{2} \mathrm{log} 2 \pi \right),
\label{eq:gpoptim}
\end{equation}

\noindent where we denote the data vector by $\bm{u}_r = \left[ \bm{u}, \dot{\bm{u}}, \Ddot{\bm{u}} \right]^T$, and gather the optimisable parameters in $\bm{\theta} = \lbrace \sigma_s^2, \sigma_{n,u}^2, \sigma_{n,\dot{u}}^2, \sigma_{n,\Ddot{u}}^2 \rbrace$. Although fully Bayesian approaches can also be adopted to derive probability distributions for each of the parameters (cf.~e.g.~\cite{tondo2023stochastic, tondoPhysicsinformedGaussianProcess2022}), this will not be investigated in this work. 

The parameter optimisation procedure in Eq.~(\ref{eq:gpoptim}) scales with a computational complexity of $\mathcal{O}(N_t^3)$ due to the matrix inversion operation, making it infeasible for very long or densely sampled measurements. Approaches to overcome this issue are either based on global approximations~\cite{hensman2013gaussian, titsias2009variational, quinonero2005unifying, chalupka2013framework, snelson2005sparse}, which summarize the dataset in a small sparse selection of data points and thus ignores finer details in the data, or in local approximations~\cite{tresp2000bayesian,hinton2002training,liu2018generalized}, which generally involves training multiple models based on subsets of data and later combining them, often leading to discontinuous predictions and overfitting. An extensive review of such methods is given in~\cite{liu2019understanding}. 

Herein, we exploit the basis functions approach for the covariance kernel creation to accelerate model training. Initially, we recall that the measurements of structural responses are generally sparse in the frequency domain since the system's mechanical admittance acts as a filter for any broadband input excitation. As a result, the original frequency vector $\bm{f}$ with size $N_f \times 1$, used to create the basis functions matrix, can be reduced to a subset $\bm{f}_r$ with size $N_{f,r} \times 1$, with $N_{f,r} \ll N_f$. The selection of frequencies in $\bm{f}_r$ can be made by filtering out regions with low spectral amplitude $\mathcal{F} \left[ {u} \right] (f)$ in the measurement data. This now yields a basis function matrix $\bm{\Psi}_i$, for $i \in \lbrace u, \dot{u}, \Ddot{u} \rbrace$, with size $N_t \times 2 N_{f,r}$, but does not change the size of the full covariance matrix $\bm{\Sigma}$, and therefore does not yet reduce the optimisation's computational complexity. However, re-writing the covariance matrix $\bm{\Sigma}$ as a function of the response basis functions $\bm{\Psi}_r$ yields

\begin{equation}
\bm{\Sigma} = \bm{\Psi}_r (\sigma_s^2 \bm{I}) \bm{\Psi}_r^T + \bm{\Sigma}_n,
\label{eq:gpbasisfull}
\end{equation}

\noindent which can be inverted using the Woodbury matrix identity~\cite{higham2002accuracy} as 

\begin{equation}
\bm{\Sigma}^{-1} = \bm{\Sigma}_n^{-1} - \bm{\Sigma}_n^{-1} \bm{\Psi}_r ((\sigma_s^2 \bm{I})^{-1} + \bm{\Psi}_r^T \bm{\Sigma}_n^{-1} \bm{\Psi}_r)^{-1} \bm{\Psi}_r^T \bm{\Sigma}_n^{-1}.
\label{eq:gpbasisinv}
\end{equation}

\noindent Since $\bm{\Sigma}_n$ and $\sigma_s^2 \bm{I}$ are diagonal matrices, they can be inverted using scalar operations only. Further, because of its size, the inversion of $((\sigma_s^2 \bm{I})^{-1} + \bm{\Psi}_r^T \bm{\Sigma}_n^{-1} \bm{\Psi}_r)$ has computational complexity of $\mathcal{O}(N_{f,r}^3)$, which is much more amenable than the na{\"i}ve direct inversion of $\bm{\Sigma}$. Therefore, filtering out irrelevant frequency components in the response signal improves training efficiency by reducing the size of matrices to be inverted. However, frequency filtering must be done carefully, as components outside the basis functions are not captured by the GP model. This means that all expected frequency components relevant to the forcing signal, whether or not they align with the structure’s natural frequencies, should be included in the basis functions. Additionally, forcing components in regions of low mechanical admittance, which do not produce significant structural responses, cannot be captured by the GP model. Finally, the determinant of $\bm{\Sigma}$, also used in Eq.~(\ref{eq:gpoptim}), can be efficiently computed using the matrix determinant lemma~\cite{harville1998matrix} by

\begin{equation}
\mathrm{det}(\bm{\Sigma}) = \mathrm{det}((\sigma_s^2 \bm{I})^{-1} + \bm{\Psi}_r^T \bm{\Sigma}_n^{-1} \bm{\Psi}_r) \ \mathrm{det}(\sigma_s^2 \bm{I}) \ \mathrm{det}( \bm{\Sigma}_n).
\label{eq:gpbasisdet}
\end{equation}

In the implementation of the likelihood optimisation, a small numerical perturbation is added to the noise covariance~$\bm{\Sigma}_n$ to improve stability in the numerical inversion operation in Eq.~\ref{eq:gpbasisinv}, which is particularly important in the case of noise-less measurements, where $\sigma_{n,i}^2 \to 0$ for $i \in \lbrace u, \dot{u}, \Ddot{u} \rbrace$. The matrix inversion operations are carried out using Cholesky decomposition~\cite{higham2002accuracy, rasmussenGaussianProcessesMachine2006}.

\subsection{Force prediction}
\label{sec:GPModelPrediction}

Once the prior model for the structural responses has been created and the free parameters have been optimised, an extension of the measurement GP can be created to represent the underlying dynamic forces. Substituting the structural modal responses in the oscillator equation (Eq.~(\ref{eq:dinoscC})) by their respective basis functions models ${\psi}_{u}$, ${\psi}_{\dot{u}}$ and ${\psi}_{\Ddot{u}}$ yields the basis function model for the forces

\begin{equation}
m {\psi}_{\Ddot{u}} + 2 m \zeta \omega_n {\psi}_{\dot{u}} + m \omega_n^2 {\psi}_{u} = {\psi}_{q},
\label{eq:basisforce}
\end{equation}

\noindent  where $m$, $\zeta$ and $\omega$ are the modal mass, damping ratio and natural frequency relative to each particular modal component in the analysis. These parameters can be estimated through system identification strategies or, if available, obtained from calibrated numerical models of the structure. Using the basis functions matrix $\bm{\Psi}_q = \psi_q (\bm{t}, \bm{f})$ and following the logic derived in Eq.~(\ref{eq:GeneralCovCration}), the prior model for the measurements can be augmented to include the forcing component by

\begin{equation}
     \begin{bmatrix} 
     \bm{u}_r\\
     \bm{q}\\
     \end{bmatrix}
     \sim \mathcal{N} \left( \begin{bmatrix} 
     \bm{0}\\
     \bm{0}\\
     \end{bmatrix}, \begin{bmatrix} 
     \bm{\Sigma} & \sigma_s^2 \bm{\Psi}_r \bm{\Psi}_q^T\\
     \sigma_s^2 \bm{\Psi}_q \bm{\Psi}_r^T & \sigma_s^2 \bm{\Psi}_q \bm{\Psi}_q^T\\
     \end{bmatrix} \right), 
\label{eq:gpwithforce}
\end{equation}

\noindent where we note that the noise covariance $\bm{\Sigma}_n$ is only a part of the measurement covariance $\bm{\Sigma}$, and therefore independent from the force model. Conditioning the force $\bm{q}$ on the measurement model and dataset $\bm{u}_r$ yields the predictive posterior distribution

\begin{equation}
p(\bm{q} | \bm{u}_r, \bm{t}, \bm{f}_r, \bm{\theta}) = \mathcal{N}(\bm{\mu}_q, \bm{\Sigma}_q),
\label{eq:postforce}
\end{equation}

\noindent with the dynamic force's mean $\bm{\mu}_q$ and covariance matrix $\bm{\Sigma}_q$ calculated by

\begin{align}
     \bm{\mu}_q &= \sigma_s^2 \bm{\Psi}_q \bm{\Psi}_r^T \bm{\Sigma}^{-1} \bm{u}_r,\label{eq:forcepredmean}\\
     \bm{\Sigma}_q &= \sigma_s^2 \bm{\Psi}_q \bm{\Psi}_q^T - \sigma_s^4 \bm{\Psi}_q \bm{\Psi}_r^T \bm{\Sigma}^{-1} \bm{\Psi}_r \bm{\Psi}_q^T. \label{eq:forcepredvar}
\end{align}

\section{Fundamental studies}
\label{sec:Fundamentalstudy}

\subsection{Prior covariance kernel properties}
\label{sec:KernelProperties}

The initial focus of this study involves examining the characteristics of the prior measurement model (see Fig.~\ref{fig:PIGP}, green box, and Sec.~\ref{sec:GPMeasurementPrior}). Herein, we investigate the effects of the frequency modulation in $\bm{f}_r$, the interplay between response amplitude and measurement noise, the model's ability to represent signals with dense frequency spectrum, and the physics-informed aspect introduced by the time derivatives linking the displacement, velocity and acceleration models.

\begin{figure}[!t]
    \centering
    \includegraphics[]{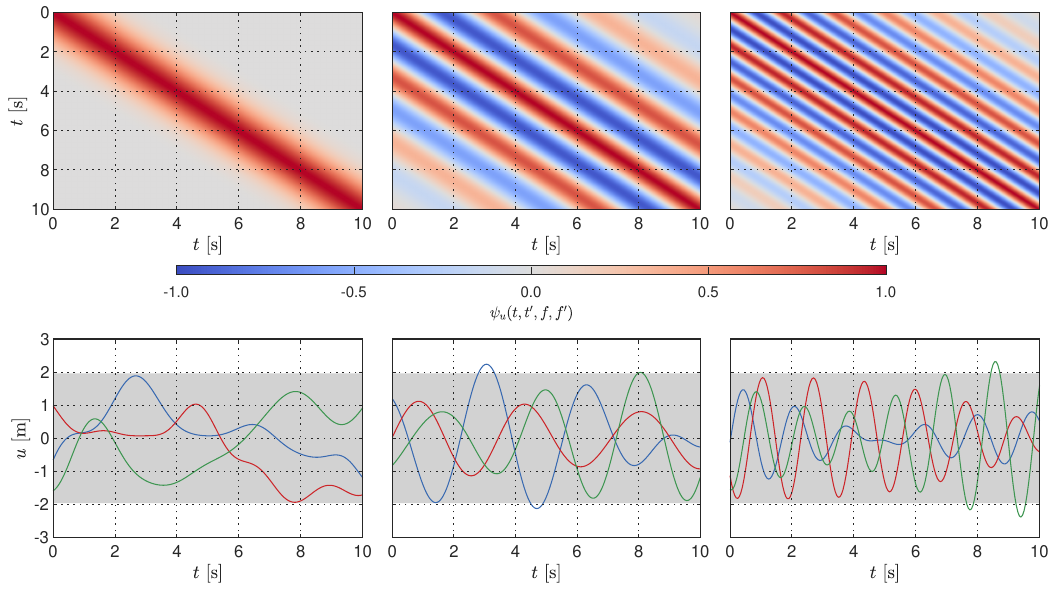}
    \caption{Top: covariance matrices $\bm{\Sigma}_{uu}$ generated with the squared exponential kernel (left), and with the basis functions model from Sec.~\ref{sec:GPMeasurementPrior} with $f=0.3$~Hz (centre) and $f=0.6$~Hz (right). Colours in the top plots indicate the covariance magnitudes, which are normalised to one. Bottom: zero-mean sample signals and 95\% confidence interval for each respective covariance matrix.}\label{fig:KernelFreqs}		
\end{figure}

We begin by studying the Gaussian process prior model for a noise-free deflection signal, $\bm{\Sigma}_{uu} = \bm{\Psi}_u \bm{\Psi}_u^T$. Unlike conventional kernel functions, such as the squared exponential (SE) kernel~\cite{rasmussenGaussianProcessesMachine2006}, the model derived using basis functions allows for the specification of the frequency content within the target signal. Figure~\ref{fig:KernelFreqs} (top) shows covariance kernels with two scenarios of single frequency content: 0.3 Hz and 0.6 Hz, alongside the SE kernel for comparison. By construction, the SE kernel characterizes similarity exclusively between proximate time instances and does not account for negative correlations. In contrast, the covariance kernels based on basis functions capture this relationship, gradually deteriorating in amplitude for time steps farther apart while accounting for the inverse relationships governed by the frequency $f$ from which they are derived. Consequently, samples generated using these covariance kernels (see Fig.~\ref{fig:KernelFreqs}, bottom) reflect these assumptions, exhibiting the anticipated frequency content, random phase, and varying amplitude over time.

\begin{figure}[!t]
    \centering
    \includegraphics[]{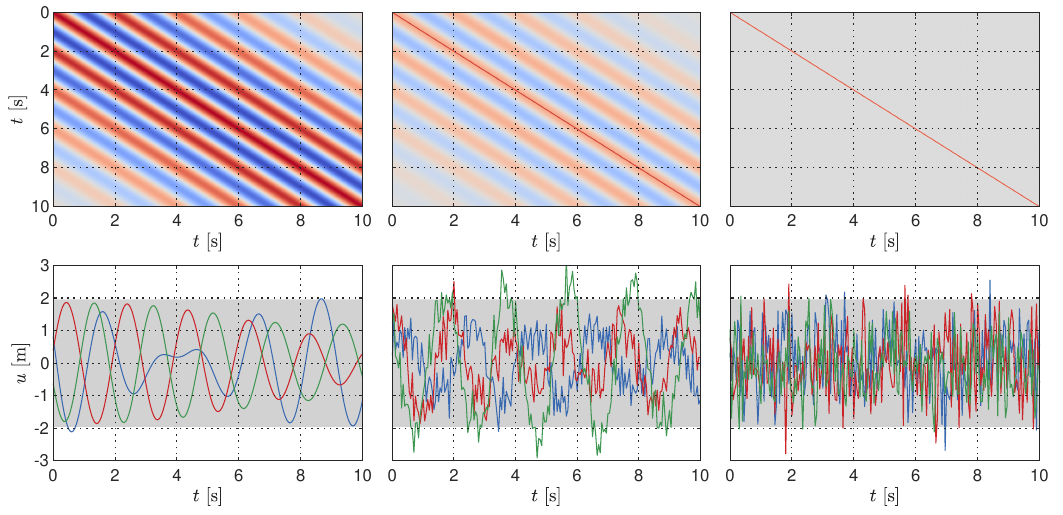}
    \caption{Influence of the ratios $r_{\sigma}$ of measurement variance to total variance in the prior model for displacements. The covariance matrices for the cases of $r_{\sigma} = 1.00$ (left), $r_{\sigma} = 0.75$ (centre) and $r_{\sigma} = 0.00$ (right) are shown on top (colours indicate the covariance magnitudes, which are normalised to one, see Fig.~\ref{fig:KernelFreqs}), with zero-mean samples generated from each respective covariance at the bottom. The shaded area in the bottom plots represents the 95\% confidence interval.}\label{fig:KernelNoise}		
\end{figure}

To account for the noise in the measurement data, a diagonal matrix of constant values $\sigma_{n,i}^2$ for $i \in \lbrace u, \dot{u}, \Ddot{u} \rbrace$ is added to the response covariance of each dataset (c.f. e.g. Eq.~(\ref{eq:NoiseDisp}) for the displacement case), following the Gaussian assumption. Due to this covariance addition, there is a relationship between the response scale parameter $\sigma_s$ and the noise scales $\sigma_{n,i}^2$. To demonstrate this interplay, using once again displacements as examples, we define the variance ratio

\begin{equation}
r_{\sigma} = \frac{\sigma_s^2}{\sigma_s^2 + \sigma_{n,u}^2},
\end{equation}

\noindent representing the percentage of the covariance matrix that is governed by the true response. Fig.~\ref{fig:KernelNoise} illustrates three cases of covariance matrices (top) with different $r_{\sigma}$ values, along with samples of time series derived from each (bottom). For $r_{\sigma} = 1.00$ (left), the model assumes no noise, being fully governed by the response covariance, resulting in a smooth time series. When $r_{\sigma} = 0.75$ (centre), the responses still predominantly govern the model's behaviour, but Gaussian noise is evident in the sampled signals. In contrast, for $r_{\sigma} = 0.00$ (right), there is no correlation between different time steps, and the samples are purely white noise with variance $\sigma_{n,u}^2$. Optimal values for $\sigma_s^2$ and $\sigma_{n,i}^2$ for all $i^{\mathrm{th}}$ datasets are identified based on the training data during the parameter optimization step, as shown in Fig.~\ref{fig:PIGP}, orange box, and Sec.~\ref{sec:GPModelTraining}.

\begin{figure}[!t]
    \centering
    \includegraphics[]{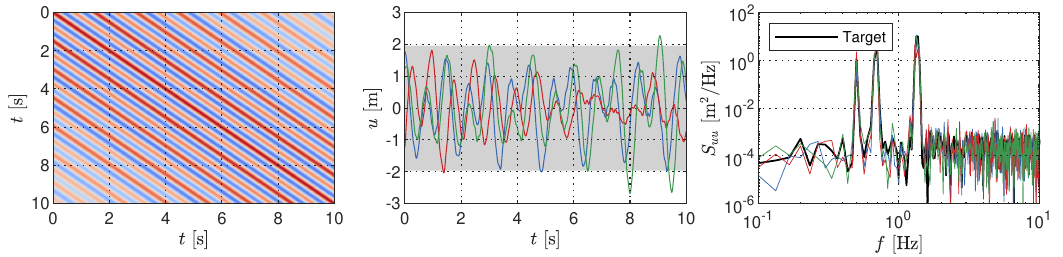}
    \caption{Prior model's response covariance matrix based on a multiple-frequency spectrum. The dense covariance matrix (left) leads to zero-mean samples containing the target frequencies (centre), as shown by their power spectral density (right). In the left plot, colours indicate the covariance magnitudes, which are normalised to one (see Fig.~\ref{fig:KernelFreqs}), and in the central plot, the shaded area represents the 95\% confidence interval.}\label{fig:KernelSpec}		
\end{figure}

As discussed in the model derivation, the basis functions are created by scaling each sine and cosine component with the amplitude of the training data's Fourier transform (see Eq.~(\ref{eq:BasisFunctU}) for the displacement case). While we have demonstrated cases with a single frequency component, the developed model is general and accommodates a densely populated frequency spectrum. To illustrate this, a synthetic target spectrum was created using three main frequency components: $f_1 = 0.50$~Hz, $f_2 = 0.75$~Hz and $f_3 = 1.35$~Hz, each with increasing amplitudes and random phase angles. The model was then created based on this spectrum, assuming a noiseless scenario. The prior covariance kernel is shown in Fig.~\ref{fig:KernelSpec} (left), displaying a rich and dense structure resulting from the multiple high-frequency components. Samples generated from this covariance (Fig.~\ref{fig:KernelSpec}, centre) also exhibit several governing frequencies and remain smooth in the time domain, as no noise prior was added to the covariance model. Furthermore, the power spectral density analysis of each sample shows a good agreement w.r.t. the target spectrum, as shown in Fig.~\ref{fig:KernelSpec} (right).

\begin{figure}[!t]
    \centering
    \includegraphics[]{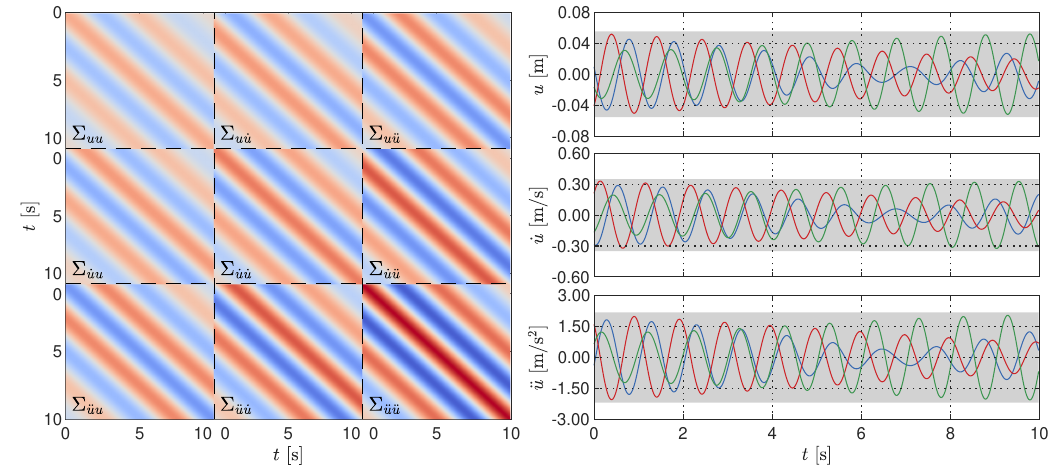}
    \caption{Left: block-covariance matrix for a heterogeneous response case, where displacements $u$, velocities $\dot{u}$ and accelerations $\Ddot{u}$ are used as training data (colours indicate the covariance magnitudes, which are normalised to one, see Fig.~\ref{fig:KernelFreqs}). Right: Time-domain samples for displacements (top), velocities (centre) and accelerations (bottom) sampled from the covariance matrix. The grey shaded area represents the signal's 95\% confidence interval.}\label{fig:KernelPhysicsInfo}	
\end{figure}

We further investigate the physical relationships between displacements, velocities, and accelerations, which are incorporated into the covariance matrices through the direct time derivatives of the basis functions (see Eqs.~(\ref{eq:BasisVel}) and (\ref{eq:BasisAcc})). Assuming a noise-free and single-frequency case, we generate the prior basis functions that yield the covariance and cross-covariance matrices. Ultimately, a full measurement covariance matrix is constructed according to Eq.~(\ref{eq:CovR}), with its structure and schematic division between block covariance matrices shown in Fig.~\ref{fig:KernelPhysicsInfo} (left). The cross-covariances $\bm{\Sigma}_{u \dot{u}}$ and $\bm{\Sigma}_{\dot{u} \Ddot{u}}$ have zero diagonals, consistent with the $\pi/2$ phase angle resulting from the time derivative. The negative diagonal observed for $\bm{\Sigma}_{u \Ddot{u}}$ reflects the $\pi$ phase angle between the two responses. This property is also evident in the signals generated from the full covariance matrix (Fig.~\ref{fig:KernelPhysicsInfo}, right), where each sample for displacements, velocities, and accelerations is linked via the respective time derivatives, analytically performed while generating the basis functions ${\psi}_{u} (t,f)$, ${\psi}_{\dot{u}} (t,f)$ and ${\psi}_{\Ddot{u}} (t,f)$.

\subsection{Force prediction accuracy of a SDOF signal}
\label{sec:ForceAccuracy}

\begin{figure}[!t]
    \centering
    \includegraphics[]{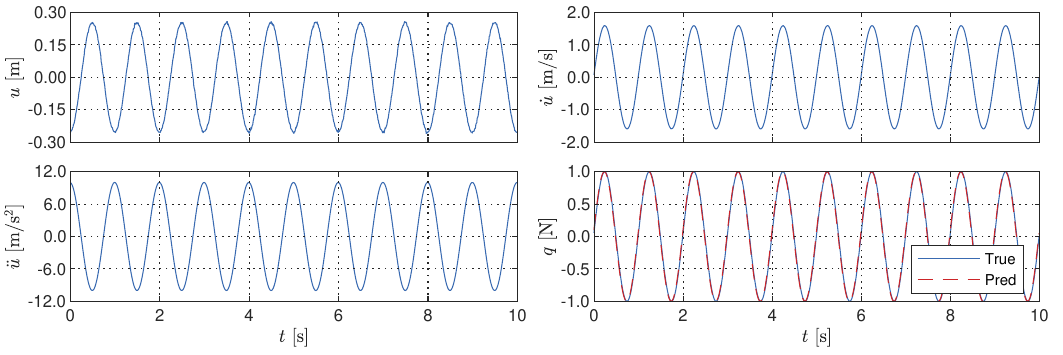}
    \caption{A sample of the force reconstruction results for a noiseless SDOF system. The model responses $u$, $\dot{u}$ and $\Ddot{u}$ are obtained numerically from the application of the true harmonic load signal. The GP model is trained using the SDOF responses, and further used to make predictions of the underlying load.}\label{fig:s001_SampleForce}		
\end{figure}

Once the response model has been created and the optimal parameters $\sigma_s$ and $\sigma_{n,i}$ for each $i^{\mathrm{th}}$ dataset have been identified, the force model can be constructed (see Fig.~\ref{fig:PIGP}, red box and Eq.~(\ref{eq:basisforce})). The joint force-measurement model can then be used to predict dynamic loads in the form of a mean function and covariance matrix, as shown in Eqs.~(\ref{eq:forcepredmean}) and (\ref{eq:forcepredvar}). At this stage, a numerical single-degree-of-freedom (SDOF) system subjected to a harmonic load is used as a basis model to investigate various properties of the measurement signals and their effects on the predicted force. An example of the SDOF responses $u$, $\dot{u}$ and $\Ddot{u}$ used to train the model, along with the true and predicted loads is shown in Fig.~\ref{fig:s001_SampleForce}. It is worth noting that, throughout this study, we assume normalised version of the mode shapes, with $\mathrm{max}(|\Phi|) = 1$, and select appropriate values for the modal mass $m$ in order to obtain the correct magnitude of the dynamic modal load. In what comes, a Monte Carlo (MC) analysis with $N = 10^4$ samples is conducted for each studied parameter, and the regressed forces are compared to the true ones using the coefficient of determination $R^2$.

\begin{figure}[!t]
    \centering
    \includegraphics[]{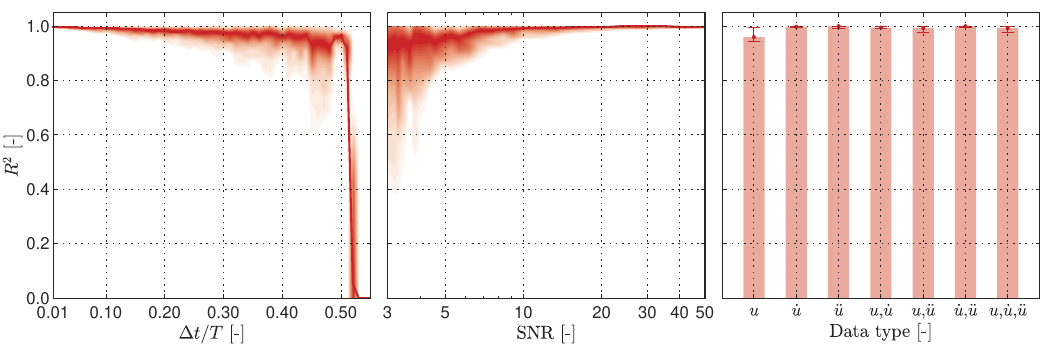}
    \caption{Coefficients of determination $R^2$ of the reconstructed force signals for different parameters of the training data. The results of sampling rate (left), signal-to-noise ratio (SNR, centre) and type of input training data (right) are shown with their respective 95\% confidence interval, according to the Monte Carlo analysis with $N=10^4$ samples.}\label{fig:FundStudiesForce}		
\end{figure}

The initial study investigates the effects of the sampling rate $f_s = 1/\Delta t$ on the quality of the predicted load. In this case, the Monte Carlo analysis is carried out for each $f_s$, with noise added to the response signals at a signal-to-noise ratio (SNR) of 15, calculated e.g. for the displacement case by

\begin{equation}
\mathrm{SNR} = \frac{\mathrm{max} |u|}{\sigma_{n,u}},
\end{equation}

\noindent where $\sigma_{n,u}$ is the standard deviation of the white noise added to the true displacement response. These contaminated signals are used as training data. The results, shown in Fig.~\ref{fig:FundStudiesForce} (left) as a function of the SDOF system's natural period of vibration $T$, indicate a substantial increase in $R^2$ values for $\Delta t/T < 0.30$. Beyond this value, not only do the average $R^2$ values drop below $0.95$, but there is also a significant increase in standard deviation, as observed from the Monte Carlo analysis for each sampling frequency case. Furthermore, following the Shannon-Nyquist sampling theorem~\cite{shannon1949communication}, predicting forces for cases where $\Delta t / T > 0.50$ is not possible since the signal to be measured cannot be accurately represented by the selected sampling rate.

The noise present in the training data can significantly impact the quality of the predicted dynamic loads. To evaluate its effects, we analyze the SDOF case using a Monte Carlo approach for SNRs ranging from 3 to 50, with a constant $\Delta t/T = 0.10$. The results (see Fig.~\ref{fig:FundStudiesForce}, centre) indicate nearly perfect force reconstruction for training signals with SNR~$>20$. For SNRs between $10$ and $20$, although the mean $R^2$ value remains above $0.95$, an increase in variance is observed in the MC analysis, indicating potential accuracy loss for specific noise distributions within the signal. For training signals with SNR~$<10$, there is significant prediction uncertainty regarding the $R^2$ coefficient, with extreme cases showing values as low as $R^2=0.40$. However, the average $R^2$ value remains above $0.90$ in all noise cases, demonstrating the good performance of the physics-informed GP model in force prediction, and guiding practical applications regarding measurement quality.

Lastly, we fix the SNR at $15$ and the sampling frequency at $\Delta t/T = 0.10$ to study the effects of the input data type, i.e. displacements, velocities, accelerations, and combinations of these, on the load prediction quality. As shown in Fig.~\ref{fig:FundStudiesForce} (right), all cases produce acceptable $R^2$ coefficients, with the worst performance in terms of mean and variance occurring when the training data consists of displacement signals only. This behaviour is expected as displacement signals attenuate high-frequency content, causing the model to interpret forces in that spectral range as measurement noise. A similar trend is observed when training data includes both velocities and displacements, although the average $R^2$ is higher and the confidence interval is much smaller. Nonetheless, the model can predict the dynamic load signal with any type of training input, including combinations of multiple heterogeneous signals.

\begin{figure}[!t]
    \centering
    \includegraphics[]{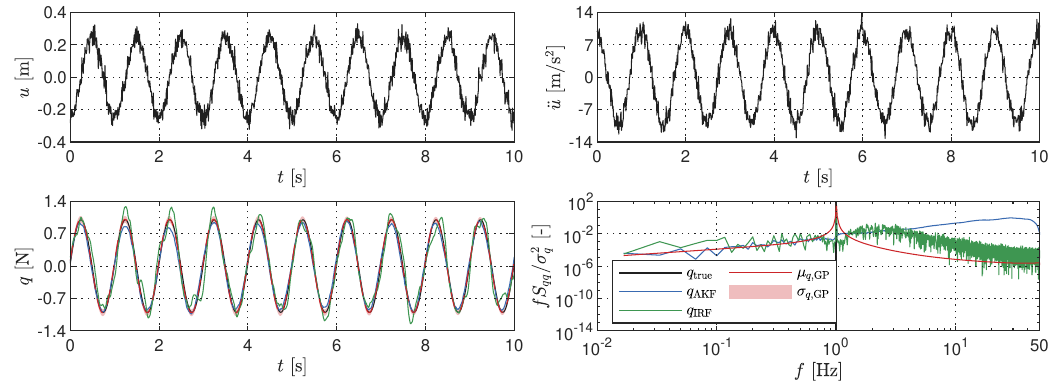}
    \caption {Accuracy comparison between different dynamic force reconstruction methods. Top: SDOF displacements (left) and accelerations (right) responses due to a harmonic force, contaminated by a white noise signal with SNR$=10$. Bottom: the reconstructed forces using the augmented Kálmán filter (AKF), impulse response function (IRF) and the proposed physics-informed Gaussian Process (GP) models, in time domain (left) and the corresponding power spectral density (right). For all models, only displacements were used as inputs.}\label{fig:s001_MethodComparison}	
\end{figure}

To further assess the performance of the physics-informed GP model, we compared its results with two established force reconstruction methods from the literature: the Augmented Kalman Filter (AKF) method~\cite{LOURENS2012446} and the Impulse Response Function (IRF) method~\cite{amiri2015derivation,draper2018numerical}. For this comparison, we used the SDOF system response to a harmonic load with frequency $f=1$~Hz, contaminated with white noise at an SNR of 10, as shown in Fig.~\ref{fig:s001_MethodComparison} (displacements at top left and accelerations at top right). Since the IRF model uses only displacement inputs, we restricted the AKF and GP models to displacements as well, fixing $\Delta t / T = 0.01$ for all. In the AKF, specific noise and uncertainty assumptions were set, including an error variance of $10^{-1}$ and noise variances for the state, augmented force, and observation equations of $10^{-15}$, $10^{-12}$, and $10^{-11}$, respectively. The IRF method employs Tikhonov regularization with a parameter $\lambda = 10^{-2}$. Both models are sensitive to these parameter choices, which were optimized using L-curve analysis~\cite{hansen1992analysis}. In contrast, the physics-informed GP model automatically balances model complexity and data fitting through its likelihood formulation (Eq.~(\ref{eq:gpoptim})), eliminating the need for additional regularization. The reconstructed force time series are shown in Fig.~\ref{fig:s001_MethodComparison} (bottom left), where the GP model shows the highest agreement with the true harmonic load, achieving an $R^2$ of 0.996, followed by the AKF model ($R^2=0.941$) and the IRF model ($R^2=0.782$). This performance is also evident in the power spectral density of the reconstructed signals (bottom left), where all models correctly identify the harmonic content at $f=1$~Hz, however, the GP model displays a smooth decay in frequencies away from $1$~Hz, while the AKF and especially the IRF exhibit higher influence from other frequency content.

\section{Experiments}
\label{sec:Applications}

The developed formulation is now applied to two case studies: a numerical simulation of a 76-story benchmark building subjected to ambient wind loads~\cite{yang2004benchmark}, and an experimental study involving a historical dynamically scaled model of the Lillebaelt Suspension Bridge, excited by a servo motor driving a rotating mass. In the first scenario, the excitation is known \textit{a priori}, enabling a direct comparison between the force reconstruction results and the true load. In the second scenario, the force is calculated from the motion prescribed to the servo motor, with the reconstruction relying on measurements obtained from the scaled model's response. We note that our focus is on the global system behavior rather than local element effects, that may arise from e.g. the application of concentrated loads. Accurately reconstructing all forces affecting system responses, although possible with the GP model, may require a high number of modes, particularly for local effects that often involve high-frequency modes, and tend to be more affected by measurement noise.

Given the increased complexity of these examples compared to the SDOF model used in the fundamental studies, we use several metrics $\mathcal{M}$ to compare the true and reconstructed forces~\cite{kavrakov2020comparison}. Specifically, we assess global and local signals features based on root mean square difference $\mathcal{M}_{\mathrm{rms}}$, correlation $\mathcal{M}_{\mathrm{c}}$, coefficient of determination $\mathcal{M}_{R^2}$, wavelet coefficients $\mathcal{M}_{\mathrm{W}}$, peak amplitudes $\mathcal{M}_{\mathrm{peak}}$, phase angles $\mathcal{M}_{\phi}$, and warped magnitude $\mathcal{M}_{\mathrm{m}}$. The advantage of these comparison metrics is that they provide a normalized similarity assessment, where values close to unity indicate similar signals, while values close to zero indicate dissimilar signals. Further details on the comparison metrics are discussed in~\ref{sec:appendix}.

\subsection{Numerical 76-story benchmark building}
\label{sec:DataParameters}

\begin{figure}[!t]
    \centering
    \includegraphics[]{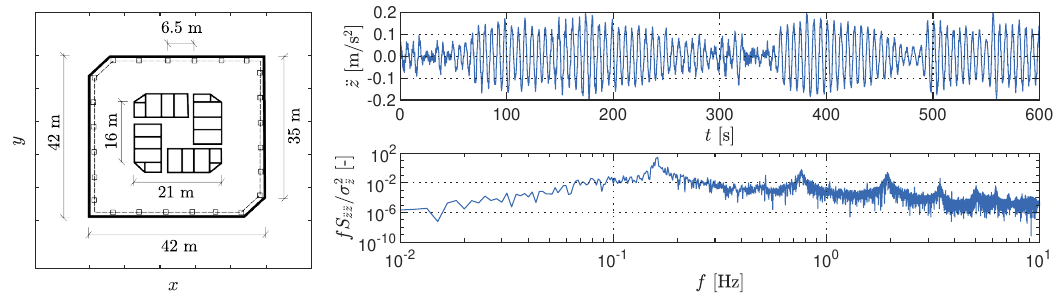}
    \caption{Left: the floor plan schematic of the benchmark building. Right: the simulated acceleration response at $H=306$~m (top) and the respective power spectral density (PSD, bottom). The responses are calculated numerically, using the load signals obtained from the wind tunnel scaled model.}\label{fig:s002_BuildingSketchResponse}		
\end{figure}

\begin{figure}[!t]
    \centering
    \includegraphics[]{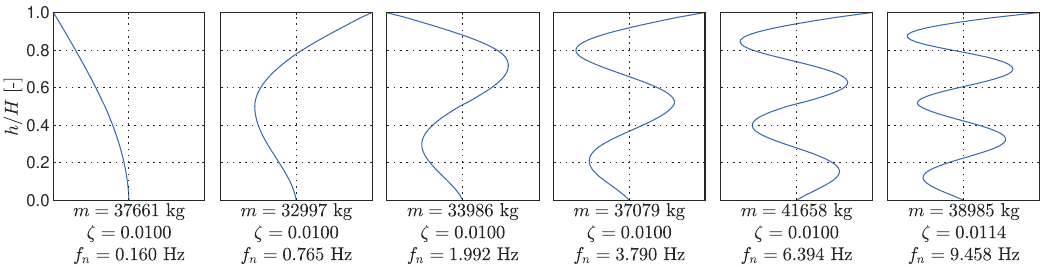}
    \caption{From left to right: first six mode shapes along the height $h$ and the respective modal masses $m$, damping ratios $\zeta$ and natural frequencies $f_n$. The 76-story building benchmark problem in Melbourne, Australia~\cite{yang2004benchmark} has a total height of $H=306$~m.}\label{fig:s002_BuildingModes}		
\end{figure}

The first application of the physics-informed Gaussian Process (GP) model for load identification involves a wind-excited 76-story concrete office tower in Melbourne, Australia. Detailed in \cite{yang2004benchmark}, this benchmark problem features a building with a total height of $H=306$~m and a square cross-section with chamfered corners (see Fig.~\ref{fig:s002_BuildingSketchResponse}, left). The building is modeled as a vertical cantilever beam with a node at each floor, resulting in 76 translational degrees of freedom (DOF), while rotational DOFs are eliminated through static condensation. The wind excitation, lasting 600 seconds, is divided into static and dynamic components, with the latter following the Davenport spectrum \cite{simiu1996wind, yang2004benchmark}. The wind load is obtained from wind tunnel testing conducted in the University of Sydney, for a rigid model of the building in a 1:400 scale, and further applied to the numerical model of the building to obtain the dynamic responses. Due to the geometric symmetry of the section and the alignment of the elastic and mass centres, no coupled lateral-torsional motion occurs. As a simplification, the benchmark considers only the along-wind component of the force. In addition, the benchmark problem allows for the control the building displacements using an active tuned mass damper. In our study, however, we use only the uncontrolled case. The acceleration response of the building's top floor and its frequency content are illustrated in Fig.~\ref{fig:s002_BuildingSketchResponse} (right). The first six normalised mode shapes and the corresponding modal mass values $m$, as well as the damping ratios $\zeta$ and natural frequencies $f_n$, are available numerically as part of the benchmark solution, as presented in Fig.~\ref{fig:s002_BuildingModes}.

In the benchmark case, the applied force is fully known, along with the system's mass, damping and stiffness matrices, used to obtain the global and modal responses of all DOFs. The known force, herein noted as $q_{\mathrm{true}}$, is used as a baseline for comparison with the forces predicted by the GP model. In this example, we first reconstruct the loads based on the first six structural modes, assuming full knowledge of the structural parameters and modal responses. Further, we contaminate the true responses with noise to simulate sensor readings, and obtain modal components by a least squares approach assuming a finite number of sensors installed in the structure. Our goal is to reproduce realistic measurement conditions in a controlled manner, and compare the force reconstruction results with the purely synthetic scenario.

\begin{figure}[!t]
    \centering
    \includegraphics[]{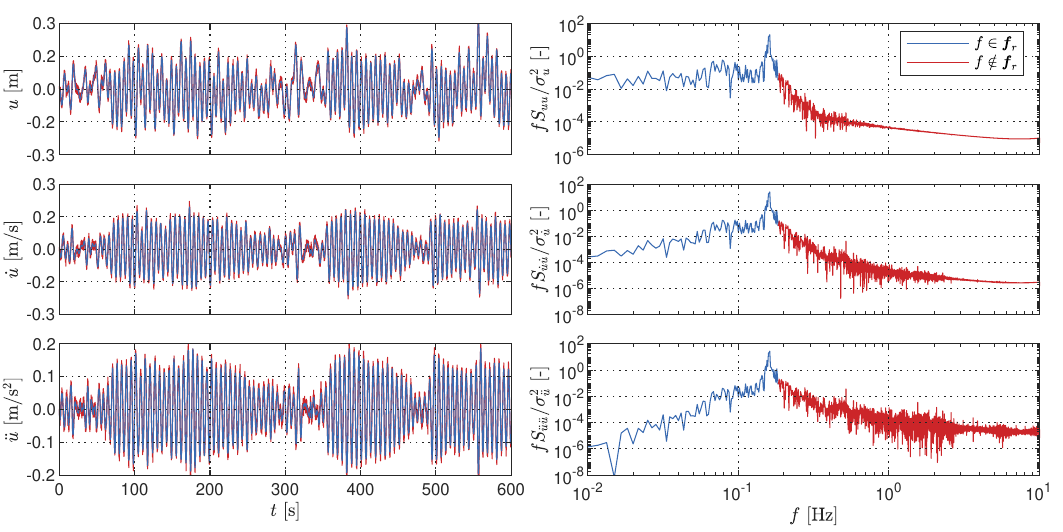}
    \caption{Left: simulated modal displacement (top), velocity (centre) and acceleration (bottom) for the benchmark building subjected to a wind excitation, considering mode 1 ($f_n = 0.160$~Hz). Right: the corresponding power spectral densities. Colours in all plots denote frequencies that are either included in the basis functions ($f \in \bm{f}_r$) or ignored by the GP model.} \label{fig:s002_ModalInputs}		
\end{figure}

We begin by identifying the first 6 aerodynamic modal loads applied to the building model based on its noise-less response, considering a dataset consisting of the three modal quantities (displacements, velocities and accelerations) sampled at $f_s = 20$~Hz. An example of the response for the first mode of vibration ($f_n = 0.160$~Hz) is shown in Fig.~\ref{fig:s002_ModalInputs} (left). As an initial step in the optimization process, we select relevant frequency components to build the response basis functions $\bm{\Psi}_r$. Different strategies are possible to accomplish this, and herein we proceed by selecting frequencies whose spectral amplitudes are higher than a cutoff limit, such that $\bm{f}_r = \lbrace f_i \in \bm{f} : |\mathcal{F}\left[ \bm{u}_r \right](f_i)| > \kappa \rbrace$, with $\kappa$ being the sum of a mean $\mu_{{\bm{u}}_r}$ and two standard deviations $\sigma_{{\bm{u}}_r}$ of $|\mathcal{F}\left[ \bm{u}_r \right] (f)|$,

\begin{equation}
    \kappa = \mu_{\bm{u}_r} + 2 \sigma_{\bm{u}_r}.    
\end{equation}

\noindent For the 6 studied mode shapes, this procedure eliminated on average $94.5\%$ of the $N_f$ frequency components contained in the response signals, significantly enhancing computational performance (c.f. Eq.~(\ref{eq:gpbasisinv})). An example of the filtered frequency content for the first mode shape is shown in Fig.~\ref{fig:s002_ModalInputs} (right).

\begin{figure}[!t]
    \centering
    \includegraphics[]{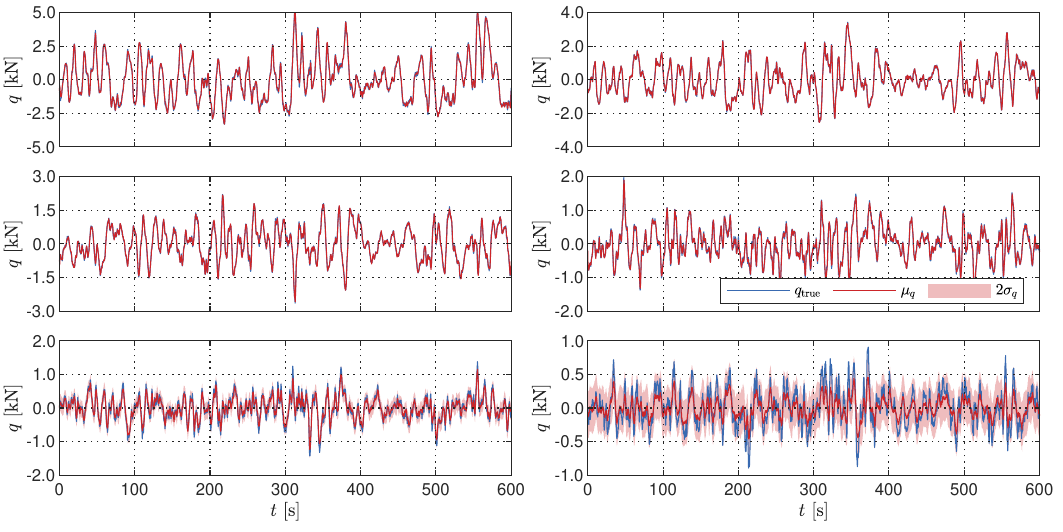}
    \caption{Benchmark modal force $q_{\mathrm{true}}$, along with mean $\mu_q$ and standard deviation $\sigma_q$ of the GP predictions (not visible for modes 1 to 4). Load signals from left to right for modes 1 and 2 (top), 3 and 4 (centre) and 5 and 6 (bottom).}\label{fig:s002_BuildingForceNoNoise}
\end{figure}

\begin{figure}[!t]
    \centering
    \includegraphics[]{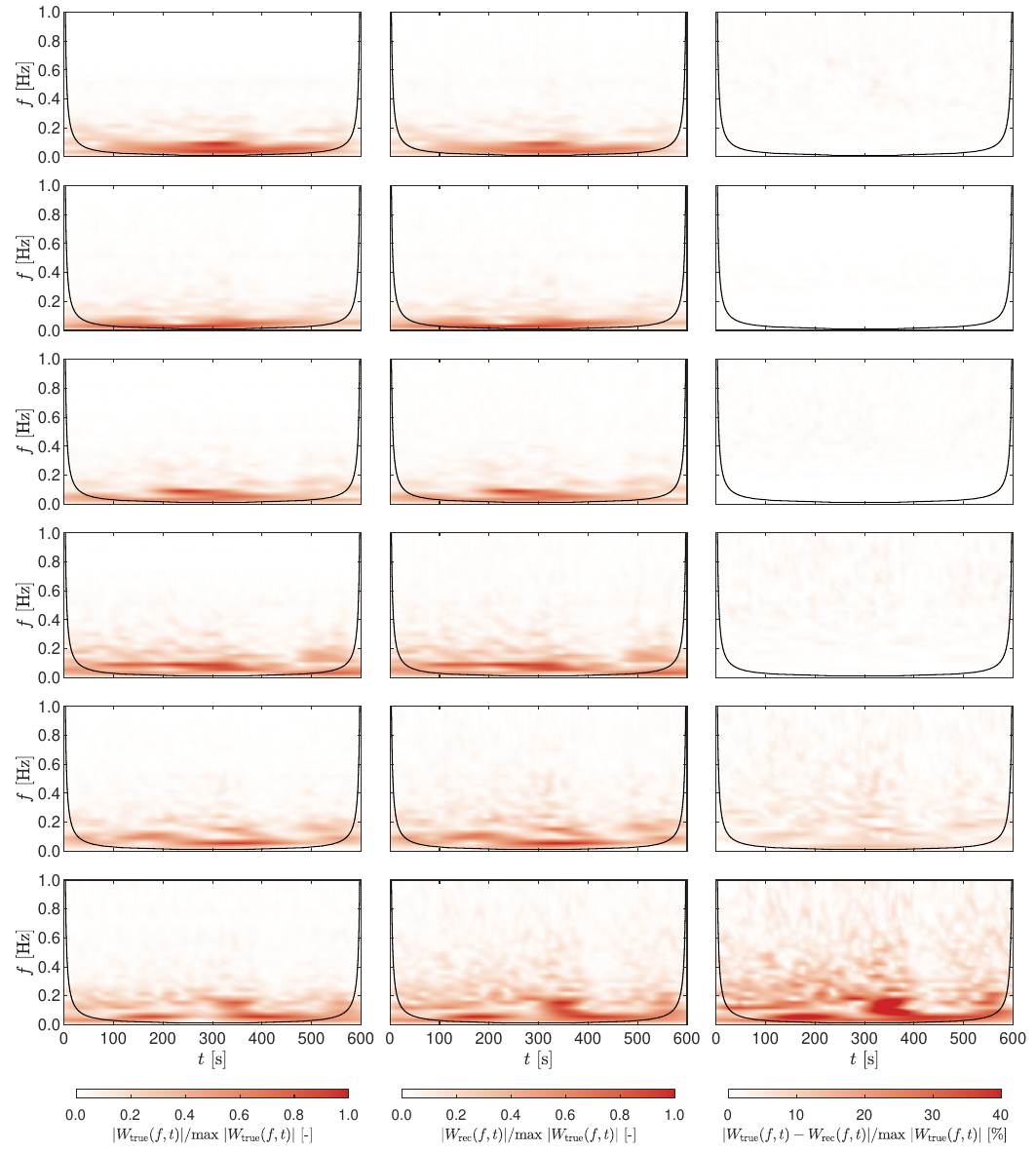}
    \caption{Morlet wavelet amplitude $|W(f,t)|$ of the true modal force (left), the reconstructed force (centre) based on noiseless responses, and their normalised difference (right), considering a central frequency $f_0 = 4$~Hz. The solid line indicates the cone of influence. From top to bottom: modes 1 to 6.} \label{fig:s002_Wavelets_Noiseless}		
\end{figure}

The force reconstruction results, shown in Fig.~\ref{fig:s002_BuildingForceNoNoise}, indicate a good overall agreement between the benchmark modal forces $q_{\mathrm{true}}$ and the mean force predictions $\mu_q$, with accuracy decreasing for higher modal numbers. The confidence interval, represented by the standard deviation of the predictions $\sigma_q$, is small compared to the mean force values for modes 1 to 4, but becomes significant for modes 5 and 6. This effect is a result of very low contribution to the response in higher modes, which is disregarded by the GP as noise. To further investigate the reconstructed signal's properties in both the time and frequency domains, we performed a wavelet analysis using a Morlet wavelet with a central frequency $f_0 = 4$~Hz. Fig.~\ref{fig:s002_Wavelets_Noiseless} shows the results for all modal cases, comparing the true and reconstructed forces, as well as the difference between them. Although the frequency content of the true modal force $q_{\mathrm{true}}$ changes over time, the physics-informed GP model accurately captures this effect, resulting in minimal errors for the first four modal cases. In mode 5, errors up to $15$\% in wavelet magnitude are observed between the true and reconstructed signals, scattered throughout the frequency content and analysis period. In mode 6, errors up to $40$\% in wavelet magnitude occur, especially during periods when particular frequencies are activated in the true force. Despite the large errors, the GP model captures the overall distribution of frequency content over time but overestimates their magnitude due to the high modal frequency ($9.458$~Hz, c.f. Fig.~\ref{fig:s002_BuildingModes}) relative to the concentrated forcing frequency content below 1 Hz. 

\begin{figure}[!t]
    \centering
    \includegraphics[]{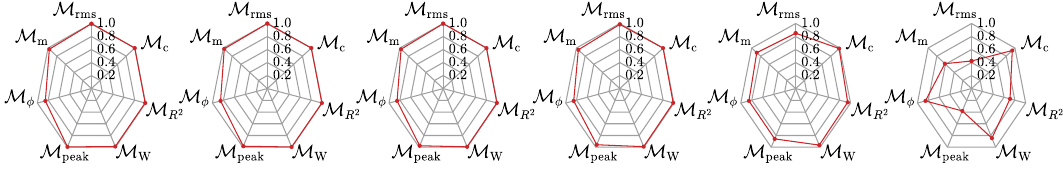}
    \caption{Comparison metrics for the force reconstruction based on the noiseless measurements of the benchmark building case. Metrics values from 0 to 1 denote worse to better results, respectively. From left to right: modes 1 to 6.} \label{fig:s002_Metrics_Noiseless_Long}		
\end{figure}

To summarize the quality of the predicted modal forces, Fig.~\ref{fig:s002_Metrics_Noiseless_Long} presents comparison metrics for each mode. As qualitatively observed in Fig.~\ref{fig:s002_BuildingForceNoNoise}, the first four modes perform well across all metrics, with values close to unity, except for the phase metric $\mathcal{M}_{\phi}$, which is approximately 0.85 in all cases. In mode 5, a reduction is observed in the root mean square $\mathcal{M}_{\mathrm{rms}}$, warped magnitude $\mathcal{M}_{\mathrm{m}}$, and peak $\mathcal{M}_{\mathrm{peak}}$ metrics, likely due to the higher modal frequency. In mode 6, significant reductions in $\mathcal{M}_{\mathrm{rms}}$, $\mathcal{M}_{\mathrm{peak}}$, and $\mathcal{M}_{\mathrm{m}}$ indicate a loss of signal magnitude throughout its duration, consistent with the behaviour observed in its reconstruction (Fig.~\ref{fig:s002_BuildingForceNoNoise}, bottom right). The wavelet metric $\mathcal{M}_{\mathrm{W}}$, while amounting to 0.82, is relatively high, indicating that the wavelet values, despite being incorrect in magnitude, align with the true force in time and frequency locations, as seen in Fig.~\ref{fig:s002_Wavelets_Noiseless}. Interestingly, the $\mathcal{M}_{\phi}$ and $\mathcal{M}_{\mathrm{c}}$ metrics remain relatively high for mode 6, indicating better performance of the model regarding phase angle and correlation retention.

\begin{figure}[!t]
    \centering
    \includegraphics[]{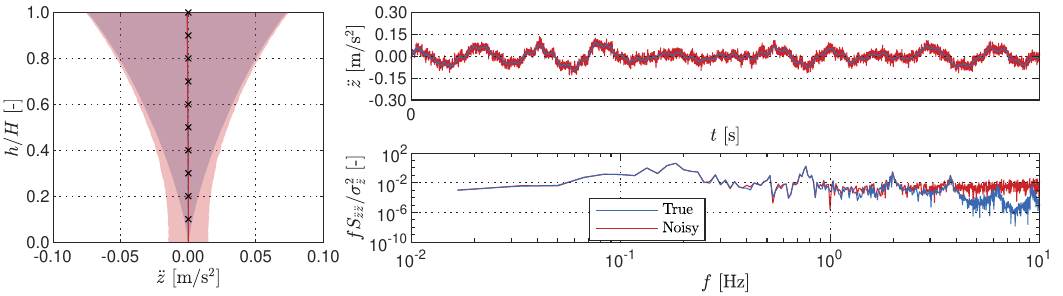}
    \caption{Benchmark building responses contaminated by white noise. Left: the mean and 2 standard deviations (shaded areas) for the true and noisy responses. Black crosses denote virtual sensor locations. Right: the true and noisy acceleration signals at floor 76 (top), and the correspondent power spectral densities (bottom).}\label{fig:s002_BuildingNoisyResponse_Long}		
\end{figure}

In real scenarios, structural responses can only be measured at discrete locations, and the collected datasets often contain significant noise. To simulate these effects, we add a white noise signal with an SNR of $20$ to the measurements, based on the building's responses at the top floor. Consequently, responses at lower building levels become much harder to identify, as the measurement signals at those locations are dominated by noise. This effect is illustrated in Fig.~\ref{fig:s002_BuildingNoisyResponse_Long} (left), which shows the mean and two standard deviations of the measurements along the height for both noise-free and noise-contaminated cases. Fig.~\ref{fig:s002_BuildingNoisyResponse_Long} (right) depicts the acceleration response at the top floor in the time domain and its respective power spectral density. To simulate discrete measurement locations, we decompose the global responses into their modal components using $N_s = 10$ measurement signals collected at equally spaced points along the height of the building (see Fig.~\ref{fig:s002_BuildingNoisyResponse_Long}, left). Assuming prior knowledge of the mode shapes at those locations, which can be obtained using system identification strategies, and considering that only the first $N_m = 4$ structural modes can be identified in the noisy signal (see Fig.~\ref{fig:s002_BuildingNoisyResponse_Long}, bottom right, and Fig.~\ref{fig:s002_BuildingModes}), the modal decomposition, e.g. for accelerations is carried out using a least squares approach by

\begin{equation}
    \Ddot{\bm{u}} = \left( \bm{\Phi}^T \bm{\Phi} \right)^{-1} \bm{\Phi}^T \Ddot{\bm{z}},
    \label{eq:modalconv}
\end{equation}

\noindent where $\bm{\Phi}$ has size $N_s \times N_m$, and the modal velocities $\dot{\bm{u}}$ and displacements $\bm{u}$ are obtained similarly from the global responses $\dot{\bm{z}}$ and $\bm{z}$. Importantly, we note that the mode shapes $\bm{\Phi}$ are magnitude-normalised to unity, and a correct modal force magnitude is achieved by selecting appropriate modal mass values $m_i$ for each $i^{\mathrm{th}}$ mode shape. In practice, correct modal mass values are difficult to obtain, and if not available, the identified dynamic loads are correct only up to a scaling factor.

\begin{figure}[!t]
    \centering
    \includegraphics[]{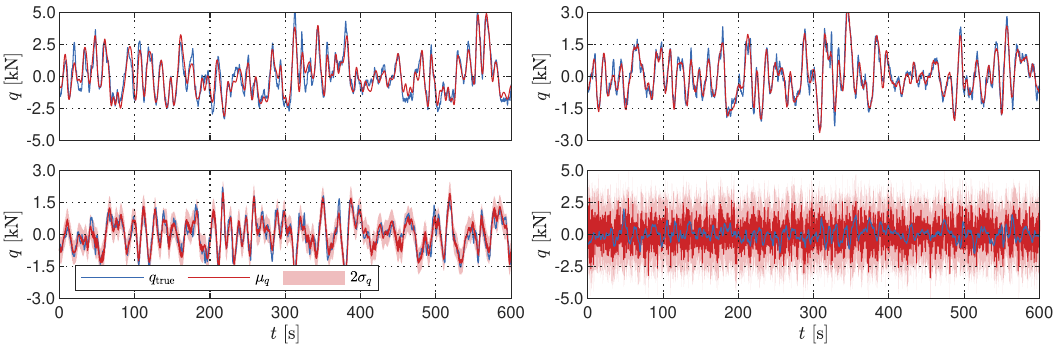}
    \caption{Reconstruction of modal forces in the benchmark building case considering noisy responses along the structure's height. True force $q_{\mathrm{true}}$ is shown along with mean $\mu_q$ and standard deviation $\sigma_q$ of the GP predictions (not visible for modes 1 to 2). Load signals from left to right for modes 1 and 2 (top), and 3 and 4 (bottom).}\label{fig:s002_BuildingForceYesNoise}		
\end{figure}

Providing the modal responses to the physics-informed GP model and identifying the optimal parameters leads to the modal force predictions shown in Fig.~\ref{fig:s002_BuildingForceYesNoise}. For the first two modes, the identification quality remains high, with only a slight reduction in mean peak amplitude and no significant prediction uncertainty in the physics-informed Gaussian process. In mode 3, an increase in standard deviation is observed, along with the presence of a high-frequency component in the forcing signal due to the noisy measurement inputs. Nonetheless, the general behaviour of the modal force is still accurately predicted. In mode 4, the influence of measurement noise becomes more pronounced, with high-frequency components from the noise visible in the time domain load signal. This results from the low relative modal contribution to the global structural behaviour, as seen in the response power spectral density in Fig.~\ref{fig:s002_BuildingNoisyResponse_Long} (bottom right).

\begin{figure}[!t]
    \centering
    \includegraphics[]{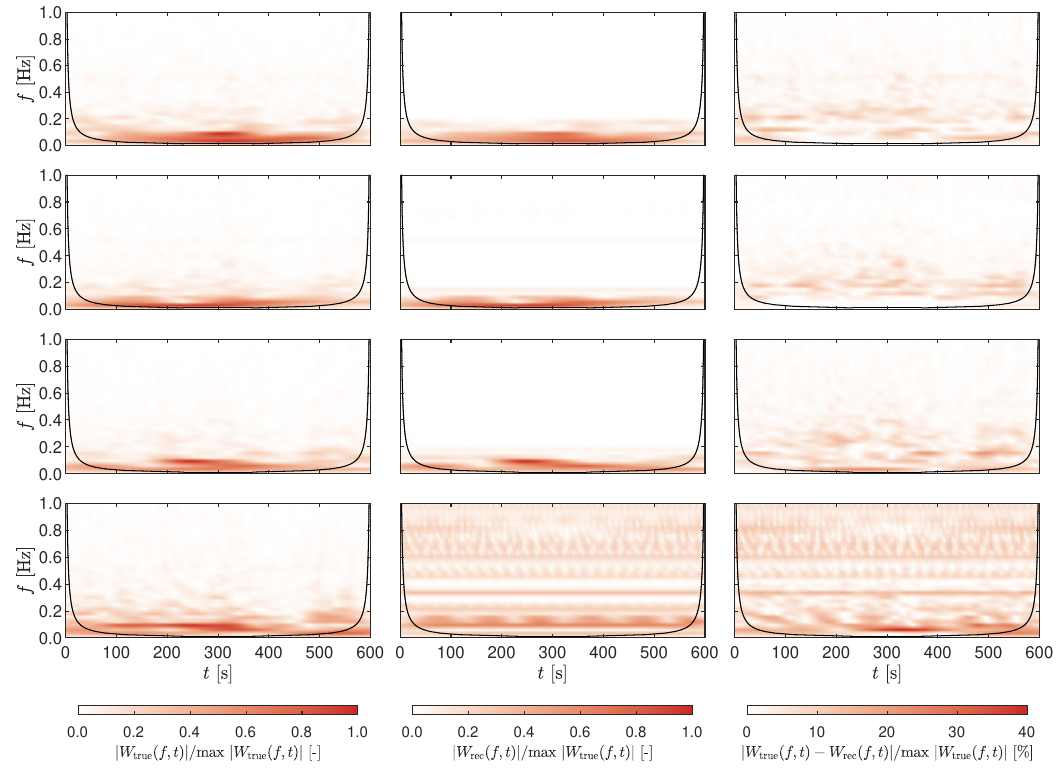}
    \caption{Morlet wavelet amplitude $|W(f,t)|$ of the true modal force (left), the reconstructed force (centre) based on noisy responses, and their normalised difference (right), considering a central frequency $f_0 = 4$~Hz. The solid line indicates the cone of influence. From top to bottom: modes 1 to 4.} \label{fig:s002_Wavelets_Noisy}		
\end{figure}

\begin{figure}[!t]
    \centering
    \includegraphics[]{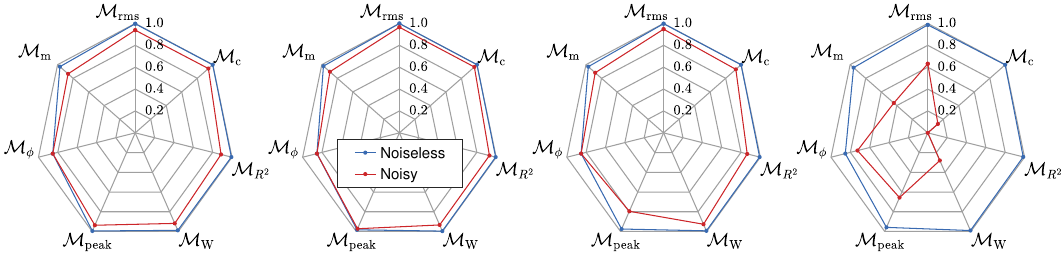}
    \caption{Comparison metrics for the force reconstruction based on the noise-contaminated measurements of the benchmark building case, with the noiseless results shown for comparison. Metrics values from 0 to 1 denote worse to better results, respectively. From left to right: modes 1 to 4.} \label{fig:s002_Metrics_Noisy_Long}		
\end{figure}

The high-frequency components in the predicted force are also evident in the wavelet magnitude (see Fig.~\ref{fig:s002_Wavelets_Noisy}). Similar to the noiseless case, the first two modal forces are accurately identified in the time-frequency plane, with only minor errors observed. However, in the third and fourth modes, significant errors in the high-frequency content are present throughout the analysis period. Specifically, in the fourth mode, the low-frequency content of the true force is not fully captured by the GP model, as indicated by the error content for $250$~s $<t<420$~s and $f<0.2$~Hz. The reconstructed modal forces are further evaluated using comparison metrics, as shown in Fig.~\ref{fig:s002_Metrics_Noisy_Long}. A slight decrease in most metrics is observed in the first three modes compared to the noiseless scenario. Interestingly, the phase metric $\mathcal{M}_{\phi}$ remains unaffected by the noise, while the peak metric $\mathcal{M}_{\mathrm{peak}}$ shows the most significant reduction, decreasing by 0.19. The results for the fourth mode reflect the time-domain signal prediction and show clear differences across all analyzed comparison metrics.

\subsection{Experimental Lilleb{\ae}lt Suspension Bridge scaled model}
\label{sec:Lillebaelt}

\subsubsection{Model and experiment description}

\begin{figure}[!t]
    \centering
    \includegraphics[]{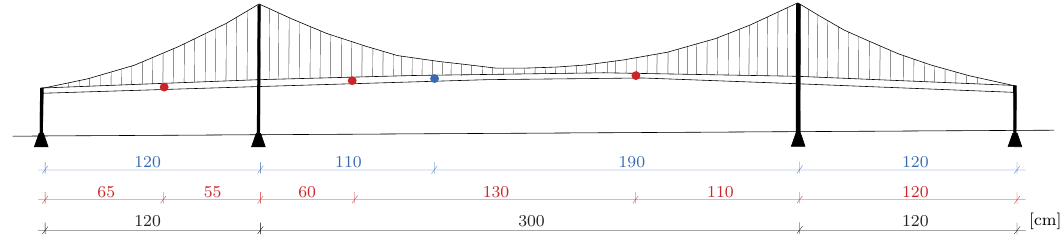} \\ \vspace{2mm}
    \includegraphics[]{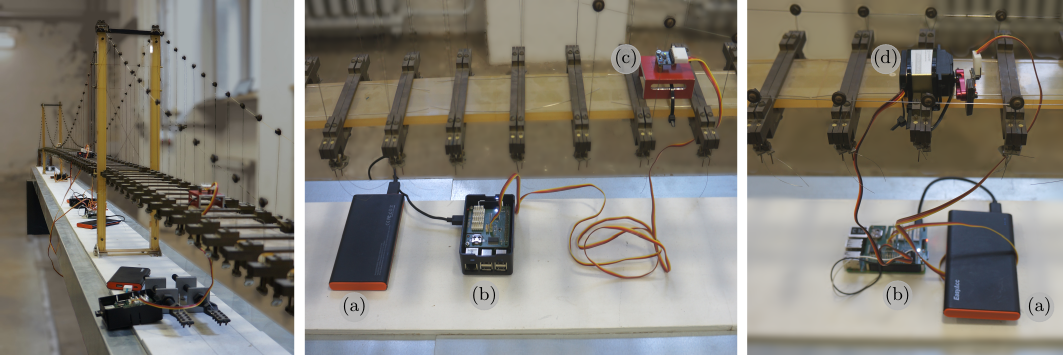}
    \caption{Top: Schematic of the 1:200 scaled model of the Lilleb{\ae}lt Bridge in Denmark, showing the servo position (blue) and measurement sensor locations (red). Dimensions in centimeters. Bottom: the scaled model of the Lilleb{\ae}lt Bridge (left), currently located at the Bauhaus University Weimar. Bottom centre: a power bank (a) supplies energy to a RaspberryPi (b), which controls the triaxial MEMS-based accelerometer/gyroscope (c) used to measure the structural responses. Bottom right: a similar framework is used to control the servo motor (d) used to excite the structure.}\label{fig:s003_PictureExperiment}		
\end{figure}

The Lilleb{\ae}lt Suspension Bridge, constructed between 1965 and 1970, was Denmark's first suspension bridge, spanning the 1200-meter Belt between the Jutland peninsula and the island of Funen. The bridge features a main span of 600~m and two side spans of 240~m each. The steel deck, 33.3~m wide, accommodates six traffic lanes and two emergency lanes, supported by hangers connected to the main suspension cables, which are anchored 180~m behind the side pylons. As part of the final design checks, a 1:200 scaled physical model (see Fig.~\ref{fig:s003_PictureExperiment}, top) was built to study the bridge's dynamic properties and roughly verify the deformations caused by traffic and constant wind loads~\cite{WenzelLillebaelt2022}. This physical scaled model is currently preserved at Bauhaus University Weimar and will be used throughout this section as a representative model for dynamic force reconstruction.

The measurement system consists of three MEMS-based triaxial accelerometers/gyroscopes of type MPU6050, controlled by Raspberry Pi microcomputers (see Fig.~\ref{fig:s003_PictureExperiment}, bottom centre)~\cite{morgenthal2019wireless, morgenthal2019effizientes}. The sampling frequency is set to 200 Hz, with a measurement range of $\pm 4$~g for the accelerometers, and $\pm 1000$~deg/s for the gyroscopes. Sensor locations, schematically shown in Fig.~\ref{fig:s003_PictureExperiment} (top), were determined using a stochastic sensor placement optimization algorithm~\cite{guestrinNearoptimalSensorPlacements2005, krauseNearOptimalSensorPlacements2008, tondo2023stochastic}. This algorithm identifies optimal locations by minimizing the entropy of a Gaussian process model with covariance defined by the mode shapes of a numerical model of the full-scale bridge. The scaled model is excited using a servo motor with an eccentric mass of $m = 24$~g attached to a lever arm of length $l = 2.9$~cm, as shown in Fig.~\ref{fig:s003_PictureExperiment} (bottom right). The motor can rotate with amplitudes of up to $\hat{\alpha} = \pm 30$~degs and has a reliable response for oscillation frequencies of up to $10$~Hz. The axis of rotation of the motor lever arm is eccentric relative to the bridge's longitudinal axis, allowing excitation of torsional modes of the deck. The forces introduced by the servo motor rotation into the bridge model, denoted throughout this section as measurement forces, are analytically calculated using rotation measurements collected by a gyroscope installed at the tip of the servo's lever arm. 

\begin{figure}[!t]
    \centering
    \includegraphics[]{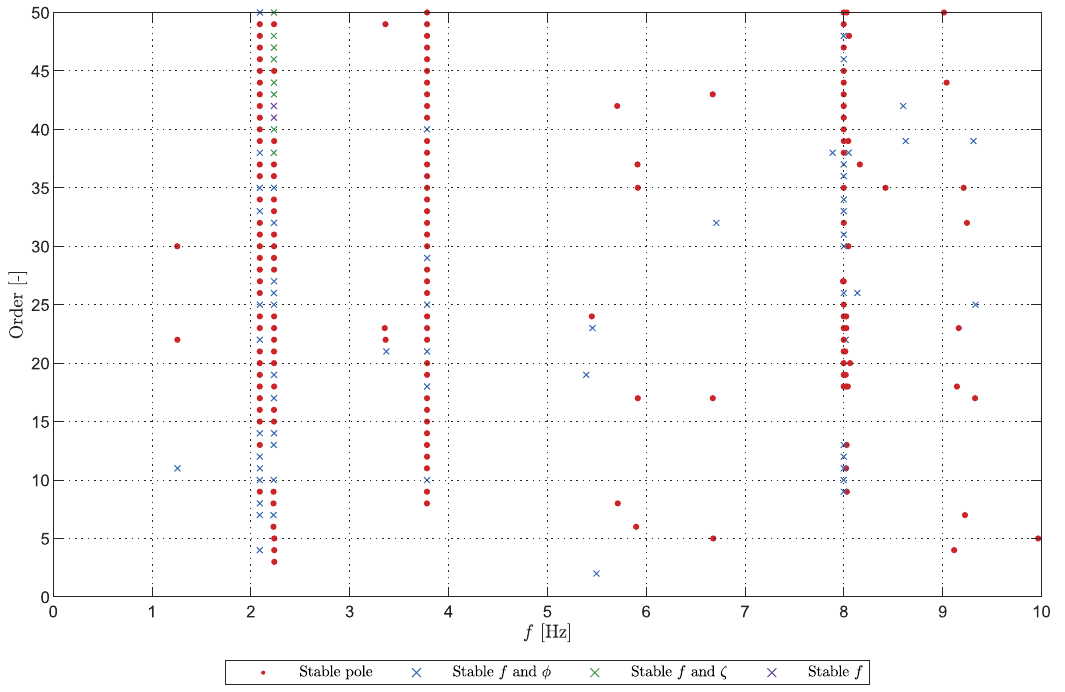}
    \caption{Stabilisation diagram of the measured response due to a white noise signal applied to the Lilleb{\ae}lt scaled model.}\label{fig:s003_StabilisationDiagram}		
\end{figure}

\begin{figure}[!t]
    \centering
    \includegraphics[]{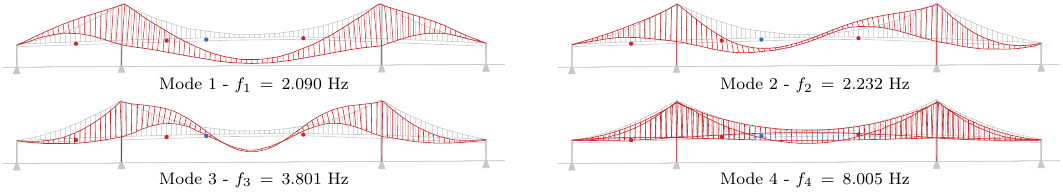}
    \caption{Schematic of the mode shapes identified in the Lilleb{\ae}lt scaled model, with the locations of the measuring sensors and servo motor shown for reference (c.f. Fig.~\ref{fig:s003_PictureExperiment}, top). It is important to note that the mode shapes used for modal decomposition in Sec.~\ref{sec:LLForceRec} are obtained experimentally from the structural response and identified via COV-SSI.}\label{fig:LBModes}		
\end{figure}

With the measurement sensors and servo motor installed on the bridge, a white noise signal is used to excite the model. The response is then processed to extract modal properties using the covariance-driven stochastic subspace identification (COV-SSI) method~\cite{peeters1999reference, peeters2000reference}. The stabilization diagram for the analysis is shown in Fig.~\ref{fig:s003_StabilisationDiagram}, where pole stability is defined by a maximum of 5\% error in frequency, damping ratio, or modal assurance criterion (MAC) for increasing system orders. Estimating the damping ratio $\zeta$ is challenging due to the assumptions about its nature and the noise effects on measurement signals. Therefore, they are estimated in a secondary step using logarithmic decrement analysis. This involves measuring free decay datasets individually for each natural frequency and mode shape, which are identified via COV-SSI. In addition, operational modal analysis techniques, such as COV-SSI, yield only arbitrarily scaled mode shapes~\cite{ewins2009modal}, meaning the underlying dynamic loads can only be determined up to a scaling factor. In practice, normalized mode shapes can be used if the correct modal masses are available, typically from calibrated finite element models or more specific system identification procedures~\cite{acunzo2018modal,hwang2015estimation}. In our application, modal masses are obtained using an experimental modal analysis technique, by applying single harmonic forces at each modal frequency identified by COV-SSI and comparing the force and response amplitudes~\cite{ewins2009modal}. The modal properties of the laboratory structure are presented in Tab.~\ref{tab:s003_simplySupportedSensorRegressionComparison}. A schematic of the mode shapes obtained from a finite element model of the full-scale structure is shown in Fig.~\ref{fig:LBModes} for reference. It is important to note that the mode shapes used for modal decomposition in Sec.~\ref{sec:LLForceRec} are obtained from the structural response and experimentally identified via COV-SSI.

\begin{table}[!t] 
	\centering
	\tabcolsep=0pt%
	\caption{Modal masses $m$, damping ratios $\zeta$ and frequencies $f$ of the scaled model of the Lilleb{\ae}lt Bridge.}		
	\begin{tabular*}{1\textwidth}{@{\extracolsep{\fill}}>{}c >{}c >{}c >{}c >{}c}
		\toprule
		Mode number & Mode type & \multicolumn{3}{c}{{Modal properties}}\\
		& & $m$ [kg] & $\zeta$ [\%] & $f$ [Hz] \\
		\midrule
		1 & Bending & 0.113 & 0.23 & 2.090 \\
		2 & Bending & 0.378 & 0.36 & 2.232 \\
		3 & Bending & 0.355 & 0.19 & 3.801  \\
		4 & Torsional & 0.027 & 0.32 & 8.005  \\	
		\toprule
	\end{tabular*}
	\label{tab:s003_simplySupportedSensorRegressionComparison}
\end{table}

\subsubsection{Pseudo-random load reconstruction}
\label{sec:LLForceRec}

\begin{figure}[!t]
    \centering
    \includegraphics[]{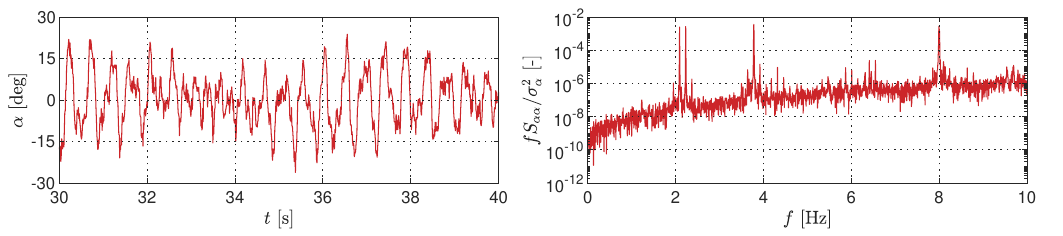} \\
    \includegraphics[]{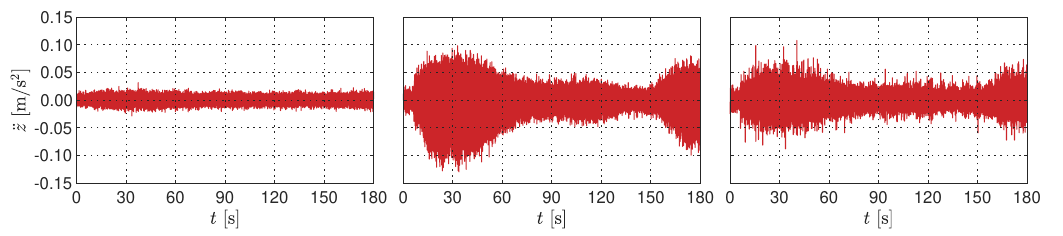}
    \caption{Top: a detailed view of the true servo lever arm rotations as measured by the installed gyroscope (left), and their corresponding power spectral densities (right). Bottom: vertical acceleration response of the scaled Lilleb{\ae}lt Bridge model, measured by the three sensors (in order, from left to right, c.f. Fig.~\ref{fig:s003_PictureExperiment}, top).} \label{fig:s003_Random_Response}		
\end{figure}

To test the force reconstruction model, we apply a pseudo-random motion to the servo motor, combining a low-level Gaussian noise and harmonic signals with frequency components within $\pm 0.1$~Hz of the scaled Lille{\ae}lt model's natural frequencies. This approach aims to excite the structure and generate measurable responses, which cannot be achieved with Gaussian noise alone due to the limitations imposed by the servo motor's attached mass and rotation amplitude and velocity. In this section, the gyroscope measurements of the rotation of the servo's lever arm (see Fig.~\ref{fig:s003_Random_Response}, top) are used to analytically calculate the forces input into the structure, which we denote as reference values $q_{\mathrm{ref}}$. The measured structural accelerations are displayed in Fig.~\ref{fig:s003_Random_Response} (bottom).

\begin{figure}[!t]
    \centering
    \includegraphics[]{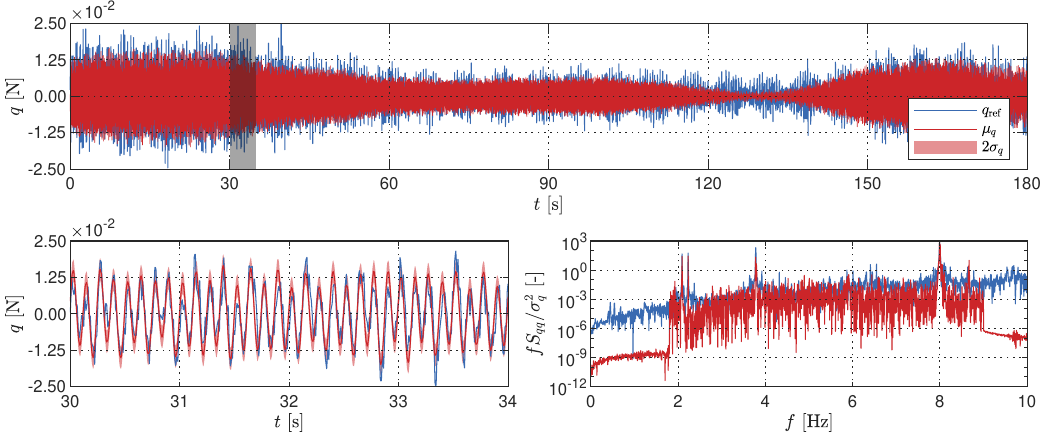}
    \caption{Top: modal force reconstruction (mode 2) of the scaled Lilleb{\ae}lt Bridge model, based on the modal decomposition of the measured responses (c.f. Fig.~\ref{fig:s003_Random_Response}). Bottom: a detailed view of the reconstructed random force (left, corresponding to the grey shaded area in the top plot), and the respective force power spectral density (right). Results for the true force (in blue) are calculated from the servo movement measured by a gyroscope installed at the motor lever arm.} \label{fig:s003_RandomTH}		
\end{figure}

The measured responses are converted into modal components using the identified mode shapes from COV-SSI and Eq.~(\ref{eq:modalconv}). These modal components, along with the corresponding modal mass, damping, and stiffness, are then input into the physics-informed Gaussian process. Given prior knowledge of the structural natural frequencies, the frequency range considered was restricted to $1.8 \leq \bm{f}_r \leq 9.0$~Hz. Figure~\ref{fig:s003_RandomTH} presents the modal force predictions for the second modal case in both the time and frequency domains, alongside a comparison with the reference forces calculated from servo movement measurements. The reconstructed forces show lower amplitudes compared to the reference forces. However, a closer examination of the time-domain results (Fig.~\ref{fig:s003_RandomTH}, bottom left) reveals generally good agreement, particularly in terms of phase angle between the two signals. In the frequency domain (Fig.~\ref{fig:s003_RandomTH}, bottom right), the GP model accurately captures peaks in relevant harmonics and does not exhibit significant components outside the specified frequency range. The absence of frequencies beyond the interest range may explain the observed reduction in force amplitude in the time domain.

\begin{figure}[!t]
    \centering
    \includegraphics[]{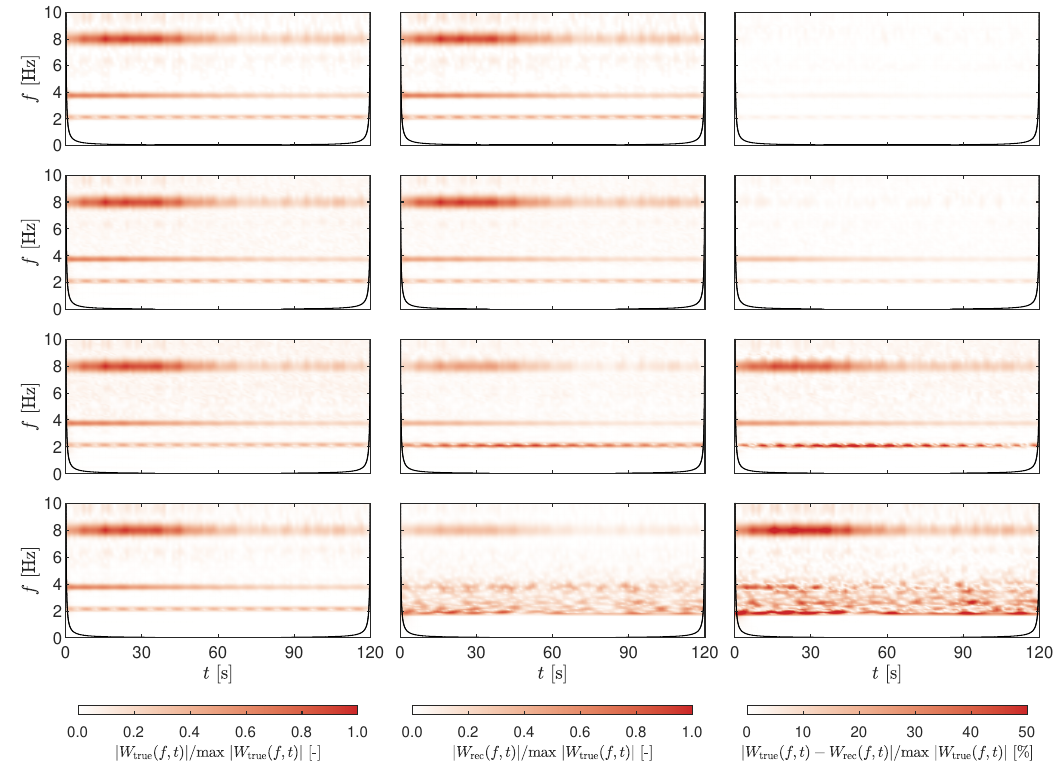}
    \caption{Morlet wavelet amplitude $|W(f,t)|$ of the reference modal force as calculated from the servo motor's measurements (left), the reconstructed force from the measured responses (centre), and their normalised difference (right), considering a central frequency $f_0 = 4$~Hz. The solid line indicates the cone of influence. From top to bottom: modes 1 to 4 (see Tab.~\ref{tab:s003_simplySupportedSensorRegressionComparison}).} \label{fig:s003_Wavelets_Random}		
\end{figure}

A detailed examination of the frequency content of the reference forces and the reconstructed forces is presented through their wavelet analysis in Fig.~\ref{fig:s003_Wavelets_Random}. As observed in previous cases, the reconstruction quality is better for the first two modal cases, where the lower structural modal stiffness results in higher modal response amplitudes and reduced influence of measurement noise. In the third mode, significant errors are noted in the frequencies governing the structure's movement. The GP model tends to underestimate the effects of high-frequency forces (e.g., $f \approx 8.0$~Hz) while overestimating the amplitudes of lower-frequency components, particularly around the first two modal frequencies ($f \approx 2.1$~Hz). For the fourth mode, the GP model underestimates the amplitude of all governing frequencies and compensates by including high-amplitude force components in the medium-to-low-frequency range ($f \leq 4.0$~Hz).

\begin{figure}[!t]
    \centering
    \includegraphics[]{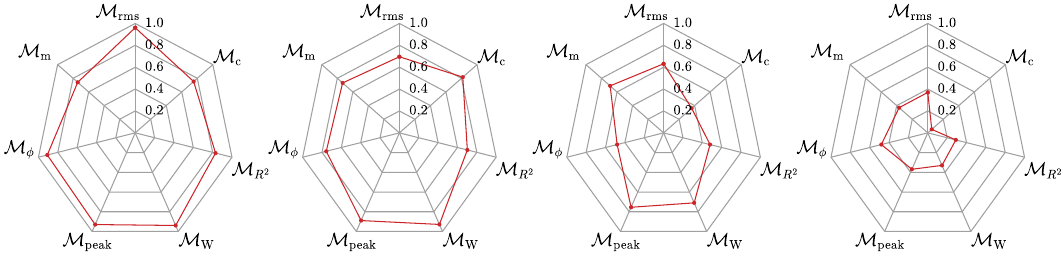}
    \caption{Comparison metrics between the reference forces calculated from the servo movement's measurements and reconstructed from the response measurements of the scaled model of the Lilleb{\ae}lt Bridge. Metrics values from 0 to 1 denote worse to better results, respectively. From left to right: modes 1 to 4 (see Tab.~\ref{tab:s003_simplySupportedSensorRegressionComparison}).} \label{fig:s003_Metrics_Random}		
\end{figure}

These observations are reflected in Fig.~\ref{fig:s003_Metrics_Random}, which shows the comparison metrics for the four modal forces relative to their reference counterparts. Generally, the first two modal cases exhibit high metric values. However, in mode 2, a decrease in $\mathcal{M}_{\mathrm{rms}}$ and $\mathcal{M}_{R^2}$ is noted, consistent with the discrepancies observed in the frequency domain (see Fig.~\ref{fig:s003_Wavelets_Random}). In the third modal case, a significant reduction in all metrics is observed, particularly in $\mathcal{M}_{R^2}$ and $\mathcal{M}_{\phi}$, indicating phase angle discrepancies and suggesting that the model struggles to replicate the measurements accurately. For mode 4, all comparison metrics fall below 0.50, indicating that the GP model fails to reconstruct the modal forces effectively, likely due to mode's low response contribution.

\section{Summary and conclusions}
\label{sec:Conclusions}

This paper introduces an efficient dynamic load reconstruction framework using a physics-informed Gaussian process built using frequency-sparse Fourier basis functions. The model's prior is built initially for structural displacements, and further extended through analytical time derivatives to develop models for velocities and accelerations, enabling joint training with multi-fidelity or heterogeneous measurement datasets.

To address the computational complexity of training GP models with large datasets, the model leverages the sparsity of structural responses in the frequency domain to filter out frequency components with low contribution to the responses, thereby reducing the number of effective basis functions. In addition, this allows for the prior specification of important target frequency ranges in the forcing signal. Matrix inversion is performed using the Woodbury matrix identity. The trained model for structural responses is then integrated with the differential equation for a harmonic oscillator, creating a dynamic load model that predicts load patterns without requiring force data during training. This approach is particularly advantageous for rare events or unusual load conditions. The model assumes linear behaviour of the structural system throughout the analysis period. In non-linear systems or in the case of additional quasi-static applied loads, this assumption can be addressed by linearizing the system around a specific equilibrium point. Alternatively, the assumption can be relaxed by shortening the prediction duration and employing system identification techniques to estimate time-varying structural parameters. In addition, other sources of uncertainty throughout the presented framework (e.g. modal mass, damping and frequency deviations, mode shape inaccuracies, sensor quantization, sensitivity and range, etc.) may also adversely affect the model prediction quality. Their individual effects on the identified dynamic modal load are propagated through the presented model, from the prior to the posterior phases, and merit further specific investigations.

The model's effectiveness was demonstrated through two case studies: a numerical model of a wind-excited 76-story benchmark building from the literature and the force reconstruction of an experimental scaled model of the Lilleb{\ae}lt Suspension Bridge, which was excited using a dedicated servo motor. Filtering irrelevant frequency content was explored, and the impact on the reconstructed forces was analyzed in both the time and frequency domains.

The physics-informed GP model exhibited good force reconstruction quality, as shown in the time-frequency plane and validated through various comparison metrics, including root mean squared values, magnitudes, phase angles, peaks, correlations, the coefficient of determination, and wavelet analysis. Generally, the model achieved better agreement between true and reconstructed forces in low-frequency regions, which are less affected by noise in MEMS-based sensors and, in our examples, have higher structural mechanical admittance values. Conversely, reconstructing higher modal contributions proved more challenging due to their lower response contribution. In general, high frequency content was captured by the GP model as part of its uncertainty range.

The developed model incorporates a self-balancing mechanism that manages the trade-off between data fitting and model complexity, eliminating the need for regularization parameters often required by similar force reconstruction algorithms. It also enables the simultaneous use of multiple datasets of structural responses during training without requiring measurement synchronization or regular sampling intervals. This model holds potential for applications in structural health monitoring, structural damage assessment, and load model validation. Future research may focus on extending the model to handle non-linear systems and non-stationary events, such as impact loads, and identifying loads associated with rare structural events, such as extreme wind conditions or specific traffic scenarios.

\section*{Acknowledgments}
IK gratefully acknowledges the support by the German Research Foundation (DFG) [Project No.  491258960], Darwin College and the Department of Engineering, University of Cambridge.

\bibliographystyle{arxiv} 
\bibliography{Man_2023_ForceRec_arXiv.bib}

\appendix
\section{Comparison metrics}
\label{sec:appendix}
Herein, we briefly introduce each of the employed comparison metrics. Complete and detailed information is given by Kavrakov et al.~\cite{kavrakov2020comparison}. In general, a comparison metric lies in the interval between 0 and 1, where unity indicates no discrepancies between two signals $x$ and $y$. It is calculated as

\begin{equation}
\mathcal{M}(x,y) = e^{-\lambda A(x,y)},
\end{equation}

\noindent where $\lambda \geq 0$ is a sensitivity parameter and $A(x,y)$ in a relative exponent that depends on each metric.

The root mean square metric $\mathcal{M}_{\mathrm{rms}}$,  is closely related to the signal energy, as it can be directly obtained from its power spectral density. Herein, its relative exponent is calculated as

\begin{equation}
A_{\mathrm{rms}} = \frac{\bigg| \sqrt{\frac{1}{T} \int_0^T \left[ x(t) \right]^2 dt} - \sqrt{\frac{1}{T} \int_0^T \left[ y(t) \right]^2 dt} \bigg|}{\sqrt{\frac{1}{T} \int_0^T \left[ x(t) \right]^2 dt}}.
\end{equation}

The correlation metric $\mathcal{M}_{\mathrm{c}}$ indicates the degree of linear correlation between the two signals as a factor or their covariance normalised by their standard deviation, and is defined by

\begin{equation}
A_{\mathrm{c}} = \frac{\mathrm{cov}(x,y)}{\sigma_x \sigma_y}.
\end{equation}

\noindent This metric is closely related to the phase metric $\mathcal{M}_{\phi}$, which accounts for mean phase discrepancies and is obtained as

\begin{equation}
A_{\phi} = \frac{\mathrm{arg max} \left[ x \star y \right]}{T_c},
\end{equation}

\noindent where $\star$ is the cross-correlation operator and $T_c$ is a normalisation time considered to be a significant delay between the signals $x$ and $y$.

The peak metric $\mathcal{M}_{\mathrm{peak}}$ measures the signal's highest value discrepancies, and its relative exponent is defined as

\begin{equation}
A_{\mathrm{peak}} = \frac{|\mathrm{max}|x| - \mathrm{max}|y| |}{\mathrm{max}|x|}.
\end{equation}

The warped magnitude metric $\mathcal{M}_{\mathrm{m}}$ determines signal discrepancies based on the dynamic time warping (DTW) algorithm, thereby considering time-localized effecs and alleviating phase shift issues. The corresponding exponents is calculated by

\begin{equation}
A_{\mathrm{m}} = \frac{\sqrt{\frac{1}{N_w} \sum_i^{N_w} \left( x_{i,w} - y_{i,w} \right)^2}}{\sqrt{\frac{1}{N_w} \sum_i^{N_w} \left( x_{i,w} \right)^2}},
\end{equation}

\noindent where $x_{j,w}$ and $y_{j,w}$ for $i = \lbrace 1, ..., N_w \rbrace$ are warped versions of the discretised signals $x$ and $y$.

The wavelet metric $\mathcal{M}_{\mathrm{W}}$ quantifies differences in the time-frequency plane of a signal in a localised manner, similar to the warped magnitude metric. It is obtained as

\begin{equation}
A_{\mathrm{W}} = \frac{1}{T} \int_0^T \frac{\sqrt{\int_0^\infty \left[ \frac{|W_x|}{W_{xf,\mathrm{max}}} - \frac{|W_y|}{W_{yf,\mathrm{max}}} \right]^2 df}}{\sqrt{\int_0^\infty \left[ \frac{|W_x|}{W_{xf,\mathrm{max}}} \right]^2 df}},
\end{equation}

\noindent where $W_i(t,f)$ is the wavelet coefficient for signal $i$, and $W_{if_j,\mathrm{max}} = \mathrm{max} |W_i (t,f_j)|$ is a normalization factor for each frequency component $f_j$. Finally, the coefficient of determination $\mathcal{M}_{R^2}$ metric measures the proportion of prediction error in relation to the variance of the original signal. This metric has a maximum value of 1 by definition, being calculated as

\begin{equation}
\mathcal{M}_{R^2} = 1 - \frac{\sum \left(x - y \right)^2}{\sum \left( x - \mu_x \right)^2},
\end{equation}

\noindent with $\mu_x$ being the mean of the reference signal.

\end{document}